\documentclass{article}

\usepackage[final]{neurips_2020}

\usepackage[utf8]{inputenc} % allow utf-8 input
\usepackage[T1]{fontenc}    % use 8-bit T1 fonts
\usepackage{hyperref}       % hyperlinks
\usepackage{url}            % simple URL typesetting
\usepackage{booktabs}       % professional-quality tables
\usepackage{amsfonts}       % blackboard math symbols
\usepackage{microtype}      % microtypography
\usepackage{nicefrac}

\usepackage{algorithm2e}
\makeatletter
\renewcommand{\@algocf@capt@plain}{above}% formerly {bottom}
\makeatother
\usepackage{mathtools}
\usepackage{amsthm}
\usepackage{amssymb}
\usepackage{tikz}
\usetikzlibrary{shapes.geometric, shapes}
\usepackage{pgfplots}
\usepackage{pgfplotstable}
\usepackage[cal=dutchcal, scr=boondox]{mathalfa}
\usepackage{dsfont}
\usepackage{etoolbox} % For more complex optional arguments in commands
\usepackage{enumitem}
\setlist{nosep}
\usepackage{multicol}
\usepackage{multirow}

\DeclareMathOperator{\vcdim}{VCdim}

\DeclareMathOperator{\sign}{sign}
\newcommand{\ie}{\textit{i.e.\@ }}

\newcommand{\pr}[1]{\left( #1 \right)}
\newcommand{\floor}[1]{\left\lfloor #1 \right\rfloor}
\newcommand{\ceil}[1]{\left\lceil #1 \right\rceil}
\newcommand{\cb}[1]{\left\{ #1 \right\}}
\newcommand{\x}{\mathbf{x}}

\newcommand{\X}{\mathcal{X}}

\newcommand{\abs}[1]{\left| #1 \right|}
\newcommand{\A}{\mathcal{A}}

\renewcommand{\P}{\mathcal{P}}
\newcommand{\Q}{\mathcal{Q}}
\newcommand{\R}{\mathcal{R}}

\newcommand{\Ll}{L_{T_l}}
\newcommand{\Lr}{L_{T_r}}

\newcommand{\parti}[1]{\bar{#1}}
\newcommand{\plambda}{\parti{\lambda}}

\newcommand{\eqdef}{
    \overset{{\mbox{\tiny \textup{def}}}}{=}
}

\newcommand{\floormtwo}{{\floor{\frac{m}{2}}}}

\newcommand{\smallstirling}[2]{\ensuremath{{\begin{Bsmallmatrix}
  #1 \\ #2
\end{Bsmallmatrix}}}}

\newtheorem{theorem}{Theorem}
\newtheorem{definition}[theorem]{Definition}
\newtheorem{lemma}[theorem]{Lemma}
\newtheorem{corollary}[theorem]{Corollary}
\newtheorem{proposition}[theorem]{Proposition}

\definecolor{color2}{rgb}{0.5,0.7,0.9}
\definecolor{color1}{rgb}{0.15,0.35,0.75}
\definecolor{color3}{rgb}{0.8,0.2,0.3}

\title{Decision trees as partitioning machines to characterize their generalization properties}
\author{%
   Jean-Samuel Leboeuf \\
   Department of Computer Science and Software Engineering \\
   Universit\'e Laval, Qu\'ebec, QC, Canada \\
   \texttt{jean-samuel.leboeuf.1@ulaval.ca} \\
  \AND
   Frédéric LeBlanc \\
   Department of Mathematics and Statistics \\
   Universit\'e de Moncton, Moncton, NB, Canada \\
   \texttt{efl7151@umoncton.ca} \\
   \And
   Mario Marchand \\
   Department of Computer Science and Software Engineering \\
   Universit\'e Laval, Qu\'ebec, QC, Canada \\
   \texttt{mario.marchand@ift.ulaval.ca} \\
}

\begin{document}
\maketitle

\begin{abstract}
Decision trees are popular machine learning models that are simple to build and easy to interpret. Even though algorithms to learn decision trees date back to almost 50 years, key properties affecting their generalization error are still weakly bounded. Hence, we revisit binary decision trees on real-valued features from the perspective of partitions of the data. We introduce the notion of partitioning function, and we relate it to the growth function and to the VC dimension. Using this new concept, we are able to find the exact VC dimension of decision stumps, which is given by the largest integer $d$ such that $2\ell \ge \binom{d}{\floor{\frac{d}{2}}}$, where $\ell$ is the number of real-valued features. We provide a recursive expression to bound the partitioning functions, resulting in a upper bound on the growth function of any decision tree structure. This allows us to show that the VC dimension of a binary tree structure with $N$ internal nodes is of order $N \log(N\ell)$. Finally, we elaborate a pruning algorithm based on these results that performs better than the CART algorithm on a number of datasets, with the advantage that no cross-validation is required.
\end{abstract}

\section{Introduction}
\label{sec:introduction}

Decision trees are popular decision models that are versatile, intuitive, and thus useful in critical fields where the interpretability of a model is important.
They are particularly useful when data is limited and not organized as in a sequence or a picture.
This makes them a good alternative to deep neural networks in several cases.

Due to their expressive power, decision trees are prone to overfitting.
To handle this problem, algorithms usually make use of practical techniques such as cross-validation in the learning or the pruning step.
Unfortunately, cross-validation increases the running time of the learning algorithm and impairs the generalization of the tree when the number of training examples is small.

As an alternative, one can use learning algorithms based on generalization bounds.
Indeed, this approach has proven its value in the work of \cite{drouin2019interpretable}, where decision trees were learned on a genomic dataset with success by optimizing a sample-compression-based bound.
Such bounds guarantee that the true risk is bounded asymptotically with high probability (ignoring logarithmic terms) in $\widetilde{O}(\frac{k+d}{m-d})$, where $k$ is the number of errors made by the tree, $d$ is the size of a compressed sample and $m$ is the size of the initial dataset \citep{ms-05}.

Relative deviation bounds based on the VC dimension \citep{vapnik1998statistical, shawe1998structural} are even tighter: in $\widetilde{O}(\frac{k+d}{m})$, where $d$ is the VC dimension of the tree.
However, to be able to make use of such algorithms to learn or prune decision trees, we must have a reasonable estimate of the VC dimension of a decision tree class, given its structure.
To the best of our knowledge, there currently exists no upper bound on the VC dimension nor the growth function of binary decision trees with real-valued features that share a common structure.
The goal of this paper is to provide such bounds.

To do so, we introduce the idea of a \emph{realizable partition} and define the notion of partitioning function, a concept closely related to the growth function and the VC dimension.
We proceed to bound tightly the partitioning function of the class of decision stumps that can be constructed from a set of real-valued features, which leads us, through the use of graph theory, to find an \emph{exact} expression of its VC dimension.
To the best of our knowledge, this was previously unknown. 
We then extend our bound of the partitioning function to general binary decision tree structures, from which we derive the asymptotic behavior of the VC dimension of a tree with $N$ internal nodes.
Finally, we show how these results can have practical implications by developing a pruning algorithm based on our bounds that outperforms CART \citep{breiman1984classification} on a number of datasets.

% Moreover, with the help of graph theory, we find an \emph{exact} (but implicit) expression of the VC dimension of the class of decision stumps that can be constructed from a set of real-valued features.
% To the best of our knowledge, this was previously unknown. 
% Then, we develop a recursive expression for the partitioning functions of decision trees, which, when used with the result about decision stumps, leads to an upper bound on the VC dimension of any decision tree class.
% From this result, we find the asymptotic behavior of the VC dimension of a binary decision tree with $N$ nodes.
% Finally, we show that these results have practical implications by developing a pruning algorithm based on them which outperforms CART \citep{breiman1984classification} on a number of datasets.

\section{Related Work}
\label{sec:related_work}

For the case of binary features, \cite{simon1991vapnik} has shown that the VC dimension of binary decision trees of rank at most $r$ with $\ell$ features is given by $\sum_{i=0}^r \tbinom{\ell}{i}$.
However, the set of decision trees with rank at most $r$ includes multiple tree structures that clearly possess different individual generalization properties.
Later, \cite{mansour1997pessimistic} claimed that the VC dimension of a binary decision tree with $N$ nodes and $\ell$ \emph{binary} features is between $\Omega(N)$ and $O(N \log \ell)$, but did not provide the proof. 
Then, \cite{maimon2002improving} provided a bound on the VC dimension of oblivious decision trees, which are trees such that all the nodes of a given layer make a split on the same feature.

In 2009, \citeauthor{aslan2009calculating} proposed an exhaustive search algorithm to compute the VC dimension of decision trees with binary features. Results were obtained for all trees of height at most 4. Then, they used a regression approach to \emph{estimate} the VC dimension of a tree as a function of the number of features, the number of nodes, and the VC dimension of the left and right subtrees.

More recently, \cite{yildiz2015vc} found the exact VC dimension of the class of decision stumps (\ie trees with a single node) that can be constructed from a set of $\ell$ \emph{binary} features, which is given by $\floor{\log_2(\ell+1)} + 1$, and proved that this is a lower bound for the VC dimension of decision stumps with $\ell$ real-valued features.
They then used these expressions as base cases to develop a recursive lower bound on the VC dimension of decision trees with more than one node.
% Furthermore, they were able to extend their results to trees with more than two branches.
However, they did not provide an upper bound for the VC dimension of decision trees.

On a related topic, \cite{gey2018vapnik} found the exact VC dimension of axis-parallel cuts on $\ell$ real-valued features, which are a kind of one-sided decision stumps.
They showed that the VC dimension of this class of functions is given by the largest integer $d$ such that $\ell \geq \tbinom{d}{\floor{\frac{d}{2}}}$.
As a corollary of their result, one has that the largest integer $d$ that satisfies $2\ell \geq \tbinom{d}{\floor{\frac{d}{2}}}$ is an upper bound for the VC dimension of a decision stump, an observation they however do not make.
Using a completely different approach, we here show that this upper bound is in fact exact.
% The approach we take allows us to show that this upper bound is in fact exact.
We discuss the difference between our results and theirs in Section~\ref{ssec:decision_stumps}.

% Finally, \cite{de2019measuring} have computed the growth function in the very limited case where the feature used for each node of a tree (to make a split) is fixed.
% However, the nodes of decision trees generally choose among multiple features during learning.
% As the number of available features affects the growth function, their results cannot generally be applied.

Our work distinguishes itself from previous work by providing an upper bound for the VC dimension of any binary decision tree class on real-valued features.
% As such, we complement the work of \cite{yildiz2015vc} and provide an improvement to the VC dimension of decision stumps.
Our framework also extends to the multiclass setting, and we show that bound-based pruning algorithms are a viable alternative to CART.

\section{Definitions and notation}
\label{sec:definitions_and_notation}

Throughout this paper, each example $\x \in \X \eqdef \mathds{R}^\ell$ is a vector of $\ell$ real-valued features\footnote{
While decision trees are often used on a mixture of real-valued and categorical features, we limit the scope of this paper to real-valued features only, mainly because categorical features require a different analysis than the one presented. We discuss the obstacles that limit the direct generalization of our framework to this type of features in more detail in the conclusion.
}.
We consider the multiclass setting with labels $y \in [n]$, where $[n] \eqdef \cb{1,\dots,n}$ for some integer $n$.
Moreover, $S$ always stands for a sample of $m$ examples, and we let $x^i_j$ be the $i$-th feature of the $j$-th example of $S$.

Recall that any \emph{tree} contains two types of nodes: \emph{internal nodes}, which have one or many children, and \emph{leaves} which do not have any children.
For simplicity, internal nodes will be referred to as \emph{nodes} (in contrast to leaves).
In a \emph{decision tree}, each leaf is associated with a class label and each node is associated with a decision rule, which redirects incoming examples to its children.
Here, we are concerned with \emph{binary decision trees} where each node has exactly two children and each decision rule concerns exactly one feature.
A \emph{decision stump} is a decision tree with only one node and two leaves.
The output $t(\x)$ of a tree $t$ on an example $\x$ is defined recursively as follows. 

% \begin{definition}[Output of a binary decision tree]
% \label{def:binary_decision_tree}
% Let $t$ be a binary decision tree. Given an example $\x$, if the tree $t$ is a leaf, the output $t(\x)$ is given by the class label associated with the leaf. 
% Otherwise, if the tree $t$ is rooted at a node with a decision rule defined by a feature $i \in [\ell]$ and a threshold $\theta \in \mathds{R}$, we label the subtree which receives examples satisfying $x_i < \theta$ as the left subtree $t_l$, while we refer to the other subtree as the right subtree $t_r$.
% Therefore, the output $t(\x)$ is given by 
% \begin{equation*}
%   t(\x) \eqdef \left\{ \!\!\!
% 	  \begin{array}{cl}
% 	    t_l(\x) & \text{if } x_i < \theta\\
% 	    t_r(\x) & \text{otherwise.}
% 	  \end{array} \right.
% \end{equation*} 
% \end{definition}
\begin{definition}[Output of a binary decision tree]
\label{def:binary_decision_tree}
If the tree $t$ is a leaf, the output $t(\x)$, on example $\x$, is given by the class label associated with the leaf. 
Otherwise, if the tree $t$ is rooted at a node having a left subtree $t_l$ and a right subtree $t_r$ with a decision rule defined by feature $i \in [\ell]$, threshold $\theta \in \mathds{R}$ and sign $s \in \cb{\pm 1}$, then the output $t(\x)$ is given by 
\begin{equation*}
  t(\x) \eqdef \left\{ \!\!\!
	  \begin{array}{cl}
	    t_l(\x) & \text{if } \sign (x^{i} - \theta) = s\\
	    t_r(\x) & \text{otherwise}\, ,
	  \end{array} \right.
\end{equation*}
where $\sign(u) =+1$ if $u > 0$ and $\sign(u) =-1$ otherwise. 
\end{definition}

From now on, we use $T$ to represent the class of binary decision trees with some fixed structure. In that case, the number of nodes and leaves and the underlying graph are fixed, but the parameters of the decision rules at the nodes and the class labels at the leaves are free parameters.
% \red{Devrait-on faire un commentaire sur la structure fixe, puisque l'un des arbitres avait accroché là-dessus?}

% We use extensively the notion of \emph{partition}, and provide a clear definition here.
\begin{definition}[Partition]
\label{def:partition}
Given some finite set $A$, an \emph{$a$-partition} $\parti{\alpha}(A)$ of $A$ is a set of $a\in\mathds{N}$ disjoint and non-empty subsets $\alpha_j \subseteq A$, called \emph{parts}, whose union is $A$.
% \begin{equation*}
%     \parti{\alpha}(A) \eqdef \cb{ \alpha_j : \alpha_j \subseteq A, \alpha_j \neq \varnothing \text{ and } \bigsqcup_{j=1}^a \alpha_j = A }.
% \end{equation*}
% Here, the square union symbol denotes the disjoint union of the parts $\alpha_j$.
% Furthermore, we say that $\parti{\alpha}$ is an $a$-partition if it is constituted of $a$ parts.
\end{definition}

% As a convention, we use capital roman letters for sets of integers or examples (such as $S$ and $A$), we use overlined lower case greek letters for partitions (such as $\parti{\alpha}$), and we use script roman letters for sets of partitions (such as $\P$).

\begin{definition}[Growth function]
\label{def:growth_function}
We define the \emph{growth function $\tau_H$} of a hypothesis class $H \subseteq [n]^\X$ as the largest number of distinct functions that $H$ can realize on a sample $S$ of $m$ examples, \ie
\begin{equation}
  \tau_H(m) \eqdef \max_{S:\left|S\right|=m} \left| \cb{ h|_S : h \in H} \right|,
\end{equation}
where $h|_S \eqdef (h(\x_1), h(\x_2), \dots, h(\x_m))$, for $\x_j \in S$, is the restriction of $h$ to $S$.
\end{definition}
The growth function can sometimes be hard to evaluate exactly.
Fortunately, in the binary classification setting, one can use the VC dimension to bound the growth function as it is often easier to estimate the former than the latter.
% The growth function measures the expressive power of a set of functions on $m$ available examples, and can sometimes be hard to evaluate exactly.
% This is indeed the case for decision trees. Fortunately, in the binary classification setting, one can use the VC dimension to bound the growth function as it is often easier to estimate the former than the latter.
\begin{definition}[VC dimension]
\label{def:VC dimension}
Let $H$ be a class of binary classifiers.
A sample $S=\cb{\x_1, \dots, \x_m}$ is shattered by $H$ iff all possible Boolean functions on $S$ can be realized by functions $h\in H$. 
The \textit{VC dimension} of $H$, $\vcdim H$, is defined as the maximal cardinality of a set $S$ shattered by $H$. 
In particular, the VC dimension of $H$ is the largest integer $d$ such that $\tau_H(d) = 2^d$.
\end{definition}

\section{Partitions as a framework}
\label{sec:paritions_as_a_framework}

% \subsection{Decision trees as partitioning machines}

Binary decision trees are traditionally defined as in Section~\ref{sec:definitions_and_notation}.
However, it is useful to represent decision trees as some kind of ``partitioning machines''.
Indeed, consider a set $S$ of examples that is sieved through some tree, so that all examples are distributed among the leaves.
Then, setting aside the labels, the set of non-empty leaves exactly satisfies the definition of a partition of $S$.
However, when the leaves are labelled, if some leaves have the same label, we take the union of the identically labelled leaves to form a single part.
%Hence, we end up with a partition of $S$ containing a smaller number of parts.
Since we are interested in the set of distinct $a$-partitions that a tree class can realize, we need the following definition.   
\begin{definition}[Realizable partition]
\label{def:realizable_partition}
Let $T$ be a binary decision tree class (of a fixed structure).
An $a$-partition $\parti{\alpha}(S)$ of a sample $S$ is \emph{realizable} by $T$ iff there exists some tree $t \in T$ such that
\begin{itemize}[topsep=0pt]
    \item For all parts $\alpha_j \in \parti{\alpha}(S)$, and for all examples $\x_1, \x_2 \in \alpha_j$, we have that $t(\x_1) = t(\x_2)$;
    \item For all distinct $\alpha_j, \alpha_k \in \parti{\alpha}(S)$, and for all $\x_1 \in \alpha_j, \x_2 \in \alpha_k$, we have that $t(\x_1) \neq t(\x_2)$.
\end{itemize}
\end{definition}
Hence, the set $\P^a_T(S)$ of all distinct $a$-partitions a tree class $T$ can realize on $S$ is obtained by considering all possible rules that we can use at each node of $T$ and all possible labelings in $[a]$ that we can assign to the leaves of $T$. 
%This set is thus independent of the number $n$ of classes of the classification problem. 
We can link the growth function $\tau_T(m)$ of $T$ to $|\P^a_T(S)|$ as follows. 
Given some realizable $a$-partition $\parti{\alpha}(S)$, we have $n$ choices of label for any one part, then we have $n-1$ choices for the next one, because assigning it the same label would effectively create an $(a-1)$-partition.
This process continues until no more parts or labels are left.
Therefore, for any $a$-partition with $a \le n$, one can produce $(n)_a$ distinct functions, where $(n)_a \eqdef n (n-1) \cdots (n-a+1)$ is the falling factorial.
Consequently, the growth function $\tau_T(m)$ can be written as
\begin{align}\label{eq:growth_func_partitions_set}
    % \tau_T(m) = \max_{S:\abs{S}=m} \sum_{a=1}^{L_T} (n)_a \abs{\P^a_T(S)},
    \tau_T(m) = \max_{S:\abs{S}=m} \,\sum_{a=1}^{\mathclap{\min \cb{m,n,L_T}}}\,\,\, (n)_a \abs{\P^a_T(S)},
\end{align}
where $L_T$ denotes the number of leaves of the tree class $T$ and where the sum goes up to $\min \cb{m,n,L_T}$ so that every term in the sum stays well defined.
% To deal with the case where the number of leaves is greater than the number of classes, we simply extend the definition of the falling factorial so that $(n)_a = 0$ for $a \geq n+1$.
This hints us to an important property of a tree class, that we call the \emph{partitioning functions}.

\begin{definition}[Partitioning functions]
\label{def:partitioning_function}
The \emph{$a$-partitioning function $\pi^a_T$} of a tree class $T$ is defined as the largest number of distinct $a$-partitions that $T$ can realize on a sample $S$ of $m$ examples, \ie
\begin{equation}\label{eq:def_a-partitioning_func}
  \pi^a_T(m) \eqdef \max_{S:\abs{S}=m} \abs{ \P^a_T(S) }\, .
\end{equation}
Moreover, we refer to the set of all possible $a$-partitioning functions of $T$ for all integers $a \in [L_T]$, with $L_T$ being the number of leaves of $T$, as the \emph{partitioning functions} of the tree class $T$. 
\end{definition}
%Unlike the growth function, the partitioning functions do not grow with the number of classes, \ie they are independent of $n$.
%Furthermore, 
Since the maximum of a sum is less than or equal to the sum of the maxima of its summands, we have that 
\begin{equation}\label{eq:ub_growth_func}
    \tau_T(m) \leq \sum_{a=1}^{L_T}(n)_a \pi^a_T(m).
\end{equation}
Moreover, we have equality whenever $n=2$ or $L_T=2$ since the first term of the sum of Equation~\eqref{eq:growth_func_partitions_set} is always $\abs{\P_T^1(S)}=1$ for any $S$ with $m>0$.

Having linked the partitioning functions to the growth function, we can relate them to the VC dimension in the following way.
On one hand we have that the total number of $a$-partitions that exist on a set of $m$ elements is given by the Stirling number of the second kind, denoted $\smallstirling{m}{a}$ \citep{graham1989concrete}.
In particular, for $m\geq 1$, we have that $\smallstirling{m}{1} = 1$ and $\smallstirling{m}{2} = 2^{m-1}-1$.
In the binary classification setting, each of these partitions yield exactly 2 distinct functions by labeling the parts with the two available classes.
Thus, $T$ can realize $2^m$ binary functions iff $T$ realizes every 1- and 2-partition on $S$.
On the other hand, Definition~\ref{def:VC dimension} implies that a tree $T$ shatters a sample $S$ iff it can realize all $2^m$ functions on $S$.
Therefore, since any tree class $T$ can realize the single 1-partition, we have that $T$ shatters a sample $S$ iff it realizes every 2-partition on $S$.
Hence, the VC dimension of any tree class $T$ having at least one internal node is given by
\begin{equation}\label{eq:ndim_def_partition_func}
    \vcdim T = \max \cb{d: \pi^2_T(d) = 2^{d-1} - 1}.
\end{equation}

\section{Analysis of decision trees}
\label{sec:analysis_decision_trees}

In this section, we analyze the partitioning behavior of decision trees.
First, we present an upper bound on the 2-partitioning function of decision stumps, which allows us to recover their \emph{exact} VC dimension.
Second, we extend our result to general tree classes, which leads us to find the asymptotic behavior of the VC dimension of a binary decision tree in terms of its number of internal nodes.

\subsection{The class of decision stumps}
\label{ssec:decision_stumps}

% In the same way decision trees can be defined recursively, we aim to evaluate the partitioning functions of tree classes in a recursive manner.
% Consequently, we need to explore the partitioning functions of the tree class of a single node, \ie the class of decision stumps that can be constructed from a set of $\ell$ features.
% These results will later prove useful in our investigation of general trees.

% We first find an upper bound on the number of 2-partitions a node can realize.
% Then, we prove that this upper bound is realizable under specific conditions for decision stumps.
% This finally leads us to an exact, but implicit, expression for the VC dimension of this hypothesis class.

As the class $T$ of decision stumps has only one root node and two leaves, the only non-trivial $a$-partitioning function of $T$ is $\pi^2_T(m)$, the maximum number of 2-partitions achievable on $m$ examples.
The following theorem gives a tight upper bound of this quantity.

\begin{theorem}[Upper bound on the 2-partitioning function of decision stumps]\label{thm:ub_partitioning_func_decision_stumps}
Let $T$ be the hypothesis class of decision stumps on examples of $\ell$ real-valued features. Then %and $S$ some sample of $m$ examples on $\ell$ real-valued features.
\begin{align}\label{eq:decision_stump_partitioning_function}
    \pi^2_T(m) \leq \frac{1}{2} \sum_{k=1}^{m-1} \min \cb{ 2\ell, \binom{m}{k} },
\end{align}
and this is an equality for $2\ell \leq m$, for $2\ell \geq \binom{m}{\floor{\frac{m}{2}}}$, and for $1 \leq m \leq 7$.
\end{theorem}
% \emph{Due to space constraints, all the proofs are provided in the supplementary material only}.
\begin{proof}
% The proof, presented in Appendix~\ref{app:proof_theorem_decision_stump}, is split in 4 parts: 1) the bound itself, 2) the equality for $2\ell \leq m$, 3) the equality for $2\ell \geq \binom{m}{\floor{\frac{m}{2}}}$, and 4) the equality for $1 \leq m \leq 7$.
% They all rely on a permutation representation of the decision rules, and the proof part 3, which is the key to find the exact VC dimension of decision stump, makes use of graph-theoretical arguments.
The proof is presented in Appendix~\ref{app:proof_theorem_decision_stump}, and relies on a permutation representation of the decision rules as well as on graph-theoretical arguments to prove the equality for $2\ell \geq \binom{m}{\floor{\frac{m}{2}}}$.
\end{proof}
We conjecture that the bound is an equality for all $m$, but it is not clear how to show this. 

Let us compare the theorem with the \emph{trivial bound} that is often used for decision stumps. The trivial bound consists in exploiting the fact that for each available feature, a stump can realize at most $m - 1$ different $2$-partitions, which gives $\pi^2_T(m) \le \ell(m-1) = (1/2)\sum_{k=1}^{m-1} 2\ell$. This yields $\tau_T(m) \le 2 + 2\ell (m-1)$ for the growth function. Comparing the trivial bound with Theorem~\ref{thm:ub_partitioning_func_decision_stumps}, we see that the trivial bound becomes an equality for $2\ell \le m$ and becomes strictly larger than the bound of Theorem~\ref{thm:ub_partitioning_func_decision_stumps} for $2\ell > m$. Also, the trivial bound exceeds the bound of Theorem~\ref{thm:ub_partitioning_func_decision_stumps} by $\ell(m-1) + 1 - 2^{m-1}$ for $2\ell \ge \binom{m}{\lfloor m/2\rfloor}$ --- a gap which is at least 
\begin{equation*}
    \frac{1}{2}\sum_{k=1}^{m-1} \left[ \binom{m}{\lfloor \frac{m}{2}\rfloor} - \binom{m}{k}\right].
\end{equation*}
Each term of the sum being positive, the trivial bound can be \emph{much larger} than the proposed bound.

Now that we have a tight upper bound on the 2-partitioning function of decision stumps, it is straightforward to find the \emph{exact} VC dimension of decision stumps.

\begin{corollary}[VC dimension of decision stumps]
\label{thm:vcdim_stump}
Let $T$ be the hypothesis class of decision stumps on examples of $\ell$ real-valued features.
Then, the VC dimension of $T$ is implicitly given by solving for the largest integer $d$ that satisfies $\displaystyle 2\ell \geq \tbinom{d}{\floor{\frac{d}{2}}}$.
\end{corollary}

\begin{proof}
According to Equation~\eqref{eq:ndim_def_partition_func}, the VC dimension is given by the largest integer $m$ such that $\pi^2_T(m) = 2^{m-1}-1$.
Theorem~\ref{thm:ub_partitioning_func_decision_stumps} gives an upper bound on the 2-partitioning function of decision stumps.
Notice that for $2\ell \geq \binom{m}{\floor{\frac{m}{2}}}$, this theorem simplifies to $\pi^2_T(m) = 2^{m-1}-1$, while for $2\ell < \binom{m}{\floor{\frac{m}{2}}}$, it implies $\pi^2_T(m) < 2^{m-1}-1$. Since $\binom{m}{\floor{\frac{m}{2}}}$ is a strictly increasing function of $m$, the largest integer $m$ such that $\pi^2_T(m) = 2^{m-1}-1$ is the largest $m$ that satisfies $2\ell \geq \binom{m}{\floor{\frac{m}{2}}}$.
\end{proof}

\noindent\textbf{Remark }
Let us mention the similarities with the result of \citet{gey2018vapnik}, where they find the VC dimension of axis-parallel cuts.
They define axis-parallel cuts as some kind of asymmetric stump, where the left leaf is always labeled~0 and the right leaf is always labeled~1.
The main difference is that the VC dimension of axis-parallel cuts is given by the largest integer $d$ that satisfies $\ell \geq \binom{d}{\floor{\frac{d}{2}}}$ (the factor 2 is absent).
Their approach is a set theoretic one, and we expect it would be hard to extend it to decision stumps, particularly for the case where $m$ is odd.
Moreover, the graph theoretic approach used here (see Appendix~\ref{app:proof_part_3_vcdim_stump}) allows us to recover a tight upper bound for the growth function (and therefore applies to the multiclass setting), while theirs does not.

\subsection{Extension to general decision tree classes}
\label{ssec:analysis_of_decision_trees}

% This section is dedicated to the study of a general binary decision tree class.
% To this end, we formalize our vision of trees as partitioning machines.
% This leads us to a useful recursive decomposition of the set of partitions realizable by a tree class.
% This decomposition is then used to find an upper bound on the partitioning functions as well as a lower bound on the VC dimension of decision tree classes.

% Let us now use Proposition~\ref{prop:c-partitions-set_decomposition_decision_trees} to provide an upper bound on the partitioning functions of binary decision trees.

We now provide an extension of Theorem~\ref{thm:ub_partitioning_func_decision_stumps} that applies to any binary decision tree class, before deriving the asymptotic behavior of the VC dimension of these classes.

\begin{theorem}[Upper bound on the $c$-partitioning function of decision trees]
\label{thm:ub_partitioning_functions_decision_trees}
Let $T$ be a binary decision tree class that can construct decision rules from $\ell$ real-valued features, and let $T_l$ and $T_r$ be the hypothesis classes of its left and right subtrees. Let $L_T$ denote the number of leaves of $T$.
Then, for $m \le L_T$, we have $ \pi^c_T(m) = \smallstirling{m}{c}$, whereas for $m > L_T$, the $c$-partitioning function must satisfy
\begin{equation}\label{eq:ub_partitioning_function_tree}
    \pi^c_T(m) \leq \pr{\frac{1}{2}}^{\delta_{lr}} \sum_{k=L_{T_l}}^{m-L_{T_r}} \!\!\min \cb{ 2\ell, \tbinom{m}{k} } \hspace{-7pt} {\sum_{\substack{1 \leq a, b \leq c \\ a + b \geq c}}} \!\!\tbinom{a}{c-b} \tbinom{b}{c-a} (a + b - c)!\; \pi^a_{T_l}(k) \pi^b_{T_r}(m-k)\, ,
\end{equation}
where $\delta_{lr} = 1$ if $T_l = T_r$, and $0$ otherwise.
\end{theorem}

The proof is provided in Appendix~\ref{app:proof_of_ub_partitioning_functions}.
% It mainly consists in applying the union bound on Equation~\eqref{eq:c-partitions-set_decomposition} and exploiting some symmetries of the quantities involved.
It relies on a recursive decomposition of $\P^c_T(S)$ exposed at the beginning of the Appendix.
Note that the inequality~\eqref{eq:ub_partitioning_function_tree} of Theorem~\ref{thm:ub_partitioning_functions_decision_trees} reduces to the inequality~\eqref{eq:decision_stump_partitioning_function} of Theorem~\ref{thm:ub_partitioning_func_decision_stumps} when $T$ is the class of decision stumps.

Theorem~\ref{thm:ub_partitioning_functions_decision_trees} can be used recursively to compute an upper bound on the VC dimension of decision trees.
Indeed, starting with $m = L_T + 1$, one can evaluate the bound on $\pi^2_T(m)$ incrementally until it is less than $2^{m-1} - 1$, according to Equation~\eqref{eq:ndim_def_partition_func}.
The algorithm is presented in Appendix~\ref{app:algorithms_bound_vcdim_trees}.%, accompanied with the computed bounds for a selected subset of decision trees shown in Figure~\ref{figure:vcdim_l=10}.

From this Theorem, one can find the asymptotic behavior of the VC dimension of a binary decision tree class on examples with real-valued features.
It is stated in the following corollary.

\begin{corollary}[Asymptotic behavior of the VC dimension]\label{coro:asymptotic_VC_tree}
Let $T$ be a class of binary decision trees with a structure containing $N$ internal nodes on examples of $\ell$ real-valued features.
Then, $\vcdim T \in O\pr{ N \log(N\ell) }$.
\end{corollary}
The proof is given in Appendix~\ref{app:proof_coro_asymptotic_vcdim} and relies on inductive arguments.
% \footnotetext{There are in fact $\text{WE}(N-1)$ non-equivalent tree classes which possess $N$ internal nodes, where $\text{WE}(n)$ is the $n$ Wedderburn-Etherington number.}

% \begin{proof}
% Letting $c=2$ in Theorem~\ref{thm:ub_partitioning_functions_decision_trees}, using the fact that $2^{-\delta_{lr}} \le 1$, $\min\cb{2\ell,\binom{m}{k}} \le 2\ell$ and $\pi^c_{T}(k) \le \pi^c_T(m)$ for $k \le m$, we have
% \begin{equation*}
%     \pi^2_T(m) \le 2\ell (m-L_T) \pr{1 + 2\pi^2_{T_l}(m) + 2\pi^2_{T_r}(m) +2\pi^2_{T_l}(m)\pi^2_{T_r}(m) }.
% \end{equation*}
% We show by induction that $\pi^2_T(m) \in O((m\ell)^N)$.
% Assume $\pi^2_T(m) \le (Cm\ell)^N$ for some constant $C\ge 1$, and let $N_l$ and $N_r$ be the number of nodes in the left and right subtrees respectively, so that $N_l+N_r=N$.
% The previous equation becomes (with $m-L_T<m$)
% \begin{align*}
%     \pi^c_T(m) &\le 2m\ell\pr{1 + 2(Cm\ell)^{N_l} + 2(Cm\ell)^{N_r} + 2(Cm\ell)^{N_l} (Cm\ell)^{N_r}}\\
%     &\le 14m\ell (Cm\ell)^{N_l+N_r},
% \end{align*}
% which proves our claim for $C \ge 14$.
% Then, Equation~\eqref{eq:ndim_def_partition_func} implies
% \begin{equation*}
%     \vcdim T \le \max \cb{ m : (Cm\ell)^N \ge 2^{m-1}-1}.
% \end{equation*}
% One can solve for the inequality $(Cm\ell)^N \ge 2^m$ instead, since this implies $(Cm\ell)^N \ge 2^{m-1}-1$ is true too.
% The Lambert $W$ function can give us an exact solution, which is $m \le - \frac{N}{\ln 2} W_{-1} \pr{ -\frac{\ln 2}{C N \ell} }$.
% Since $-W_{-1}(-z^{-1}) \in O\pr{\log z}$, we have that $\vcdim T \in O\pr{N\log(N\ell)}$.

% \end{proof}

\section{Experiments}

To demonstrate the utility of our framework, we apply our results to the task of pruning a greedily learned decision tree with a structural risk minimization approach.
We first describe the algorithm, then we carefully explain the methodology and the choices made, and finally we discuss the results.

\subsection{The pruning algorithm}

\newcommand{\Naturals}{{\mathbb{N}}}

We base our pruning algorithm on Theorem~2.3 of \cite{shawe1998structural}, which states that for any distribution $D$ over a set of $m$ examples, for any countable set of hypothesis classes $H_d$ (with growth function $\tau_{H_d}$) indexed by an integer $d$, and any distributions $p_d$ on $\mathds{N}$ and $q_k$ on $[m]$, with probability at least $1-\delta$, the true risk $R_D(h)$ of any predictor $h\in H_d$ is at most  
\begin{equation}\label{eq:shawe-taylor_bound}
    \epsilon(m, k, d, \delta) \eqdef  \frac{1}{m} \pr{ 2k + 4 \ln \pr{ \frac{4 \tau_{H_d}(2m) }{\delta q_k p_d} }}\, .
\end{equation}
Although that theorem was originally stated for binary classification and for a sequence of nested hypothesis classes $H_d$ indexed by their VC dimension, it is also valid in the multiclass setting with zero-one loss if we use the growth function directly instead of the upper bound provided by Sauer's lemma.
Furthermore, it is not necessary to have nested hypothesis classes, since the main argument of the proof uses the union bound which applies for any countable set of classes.

The goal of our pruning algorithm is to try to minimize the true risk $R_D(t)$ of a given tree $t$ by minimizing the upper bound $\epsilon$.
It goes as follows.
Given a greedily grown decision tree $t$, fixed distributions $q_k$ and $p_d$, and a fixed confidence parameter $\delta$, we compute the bound $\epsilon$ associated to this tree.
Then, for each internal node of the tree, we prune the tree by replacing the subtree rooted at this node with a leaf and we compute the bound associated with the resulting tree.
Among all such pruned trees, let $t'$ be the one that has the minimum bound value.
% Let $t'$ be the tree pruned as such that has the minimum bound value.
If the bound of $t'$ is less than or equal to the bound of $t$, we discard $t$ and we keep $t'$ instead.
We repeat this process until pruning the tree doesn't decrease the bound.
The formal version of the algorithm is presented in Algorithm~\ref{algo:prune_tree_with_bound} of Appendix~\ref{app:pruning_algorithm}.

A key distinction between our proposed algorithm and CART's cost-complexity pruning algorithm is that, for each pruning step, the cost-complexity algorithm makes the choice to prune a subtree based on local information, \ie it depends only on the performance of that subtree.
In contrast, our algorithm takes into account global information about the \emph{whole} tree via its growth function.

We would like to emphasize that the bound~\eqref{eq:shawe-taylor_bound} could not be used to prune trees prior to our work, since no upper bound on the growth function of decision trees was known. Our paper provides such a bound via Equations~\eqref{eq:ub_growth_func} and \eqref{eq:ub_partitioning_function_tree}.

\subsection{Methodology}
\label{ssec:methodology}

We benchmark our pruning algorithm on 19 datasets taken from the UCI Machine Learning Repository \citep{Dua:2019}.
We chose datasets suited to a classification task with exclusively real-valued features and no missing entries.
Furthermore, we limited ourselves to datasets with 10 or less classes, as Equation~\eqref{eq:ub_growth_func} becomes computationally expensive for a large number of classes.

These datasets do not come with a defined train/test split.
As such, we chose to randomly split each dataset so that the models are trained on 75\% of the examples and tested on the remaining 25\%.
To limit the effect of the randomness of the splits, we run each experiment 25 times and we report the mean test accuracy and the standard deviation.

We compare our pruning algorithm to CART's cost-complexity algorithm as proposed by \cite{breiman1984classification}, as it is one of the most commonly used algorithms in practice (indeed, it is the implementation of the popular \texttt{scikit-learn} Python package).
Another main reason is that it is natural to compare against the cost-complexity pruning algorithm, since it approximates the complexity of a tree via the number of leaves of the tree (which is an ad hoc educated guess), while our bounds on the growth function provide a theoretically valid quantifier of the tree's complexity.

We consider 4 models: the fully grown unpruned tree as generated by CART, the pruned tree after using the cost-complexity pruning algorithm, a modification of CART's cost-complexity pruning algorithm inspired by our work, and our pruning algorithm.

The first model we consider is the greedily learned tree, grown using the Gini index until the tree has 100\% classification accuracy on the training set or reaches 40 leaves.
We impose this limit since the computation times for pruning trees become prohibitive for a large number of leaves.
We expect that this constraint does not affect results significantly since all three pruning algorithms considered reduce the number of leaves well below 40.

The second model is the CART tree, which prunes the tree from the first model according to chapter 3 of \cite{breiman1984classification}.
The idea is to assume that the true risk of a tree can be approximated via its empirical risk by adding a complexity term of the form $\alpha L_T$ to it, where $\alpha$ is a constant and $L_T$ is the number of leaves of the tree $T$.
We did a 10-fold cross-validation on the training set to find $\alpha$.

The third model is a modification to CART's cost-complexity algorithm, where instead of assuming that the excess risk of a tree is controlled solely by the number of leaves (as in the CART algorithm), we suppose that the dependence is of the form $\frac{d}{m} \log \frac{m}{d}$, where $d = L_T \log ( L_T \ell )$, $m$ is the number of examples and $\ell$ is the number of features.
The form of the dependence is inspired by the form of bound~\eqref{eq:shawe-taylor_bound}, replacing the growth function by the approximation of Sauer's lemma and using the dependence of Corollary~\ref{coro:asymptotic_VC_tree} for the VC dimension. The rest of the algorithm is then identical to CART.

Finally, the fourth model is the one proposed in the previous section.
As parameters, we fixed $\delta = 0.05$ for all experiments.
The choices of distributions $p_d$ and $q_k$ are arbitrary and should reflect our prior knowledge of the problem.
We would like $p_d$ to go to $0$ slowly as $d$ grows in order not to penalize large trees too severely.
As we are working in a multiclass setting, we cannot use the VC dimension to index the hypothesis classes.
Instead, as an approximation to the complexity index of a tree, we use the number of leaves, and we give the same probability $p_d$ to every tree with the same number of leaves.
We thus choose to let $p_d = \frac{6}{\pi^2 L_{T_d}^2} \frac{1}{\text{WE}(L_{T_d})}$, where $\text{WE}(L_T)$ denotes the $L_T$-th Wedderburn-Etherington number \citep{bona2015handbook}, which counts the number of structurally different binary trees with $L_T$ leaves.

We observed that, in the bound~\eqref{eq:shawe-taylor_bound}, the penalty accorded to the complexity of the tree is disproportionately larger that the penalty accorded to the number of errors.
This is because much of the looseness of the bound comes from the growth function.
Indeed, it is already an upper bound for the annealed entropy, and our bound of the growth function adds even more looseness on top of that.
The distribution $q_k$ offers us a chance to compensate this fact by introducing a large penalty for the number of errors $k$.
We chose $q_k$ of the form $(1-r) r^{k}$ for some $r < 1$, such that $\sum_k q_k$ is a geometric series summing to $1$.
We made a 5-fold cross-validation of $r$ on a single dataset and we stuck with this value of $r$ for all others.
We tried inverse powers of $2$ for $r$ and we took the geometric mean of $10$ draws as the final value.
The Wine dataset from the UCI Machine Learning Repository \citep{Dua:2019} gave a value of $r = 2^{-13.7} \approx \frac{1}{13308}$.
This choice makes the value of the bound $\epsilon$ larger; however, it allows to correct the gap between the complexity dependence and the dependence of the bound on the number of errors, which gives better results in practice.

When running the experiments, we observed that Equation~\eqref{eq:ub_partitioning_function_tree} was computationally too expensive to be used directly because of the sum over $k$.
Hence, we used the following upper bound instead
\begin{equation*}
    \pi^c_T(m) \leq \pr{\frac{1}{2}}^{\delta_{lr}} \!\!\!\! (m-L_T) \, 2\ell \!\! {\sum_{\substack{1 \leq a, b \leq c \\ a + b \geq c}}} \!\!\tbinom{a}{c-b} \tbinom{b}{c-a} (a + b - c)!\; \pi^a_{T_l}(m-L_{T_r}) \pi^b_{T_r}(m-L_{T_l})\, ,
\end{equation*}
which simply replaces the sum over $k$ by $m-L_T$ times the greatest term of the sum.
This modified expression was much faster to compute and had only a small impact on the bound $\epsilon$ because of the logarithmic dependence on the growth function.
It is straightforward to modify Algorithm~\ref{algo:partition_func_upper_bound} of Appendix~\ref{app:algorithms_bound_vcdim_trees} to compute this looser bound.

All experiments were done in pure Python.
The source code used in the experiments and to produce the tables is freely available at the address \url{https://github.com/jsleb333/paper-decision-trees-as-partitioning-machines}.

\subsection{Results and discussion}

Table~\ref{table:results} presents the results of the four models we tested.
The column ``Original'' corresponds to the unpruned tree, the ``CART'' column is the original tree pruned with the cost-complexity pruning algorithm, ``M-CART'' is the
modified CART algorithm with the complexity dependencies changed to reflect our findings and the ``Ours'' column is the original tree pruned with Shawe-Taylor's bound.
More statistics about the models and the datasets used are gathered in Appendix~\ref{app:experiments}.

\begin{table}[h!]
\centering
\caption{Mean test accuracy and standard deviation on 25 random splits of 19 datasets taken from the UCI Machine Learning Repository \citep{Dua:2019}. In parenthesis is the total number of examples followed by the number of classes of the dataset. The best performances up to a $0.0025$ accuracy gap are highlighted in bold.}
\label{table:results}
\vspace{6pt}
\small
\begin{tabular}{l@{\hspace{6pt}}c@{\hspace{6pt}}c@{\hspace{6pt}}c@{\hspace{6pt}}c}
\toprule
\multirow{2}{*}[-3pt]{Dataset} & \multicolumn{4}{c}{Model}\\
\cmidrule{2-5}
 & Original & CART & M-CART & Ours\\
\cmidrule{1-5}
BCWD\textsuperscript{a} (569, 2) & $0.928 \pm 0.024$ & $0.923 \pm 0.027$ & $0.930 \pm 0.017$ & $\mathbf{\mathbf{0.942 \pm 0.022}}$\\
Cardiotocography 10 (2126, 10) & $\mathbf{0.566 \pm 0.023}$ & $0.562 \pm 0.023$ & $\mathbf{0.566 \pm 0.024}$ & $\mathbf{\mathbf{0.567 \pm 0.022}}$\\
CMSC\textsuperscript{b} (540, 2) & $0.903 \pm 0.024$ & $\mathbf{0.920 \pm 0.021}$ & $\mathbf{\mathbf{0.922 \pm 0.017}}$ & $\mathbf{0.921 \pm 0.014}$\\
CBS\textsuperscript{c} (208, 2) & $\mathbf{\mathbf{0.727 \pm 0.061}}$ & $0.702 \pm 0.054$ & $0.695 \pm 0.084$ & $0.724 \pm 0.053$\\
DRD\textsuperscript{d} (1151, 2) & $0.613 \pm 0.027$ & $0.576 \pm 0.044$ & $0.602 \pm 0.040$ & $\mathbf{\mathbf{0.622 \pm 0.023}}$\\
Fertility (100, 2) & $0.790 \pm 0.060$ & $\mathbf{\mathbf{0.878 \pm 0.051}}$ & $\mathbf{0.878 \pm 0.051}$ & $0.866 \pm 0.056$\\
Habermans Survival (306, 2) & $0.660 \pm 0.062$ & $\mathbf{\mathbf{0.746 \pm 0.043}}$ & $0.721 \pm 0.043$ & $0.719 \pm 0.043$\\
Image Segmentation (210, 7) & $\mathbf{\mathbf{0.862 \pm 0.048}}$ & $0.814 \pm 0.144$ & $0.844 \pm 0.050$ & $0.858 \pm 0.050$\\
Ionosphere (351, 2) & $\mathbf{0.891 \pm 0.035}$ & $0.772 \pm 0.108$ & $0.867 \pm 0.057$ & $\mathbf{\mathbf{0.892 \pm 0.032}}$\\
Iris (150, 3) & $0.933 \pm 0.030$ & $0.860 \pm 0.139$ & $0.838 \pm 0.158$ & $\mathbf{\mathbf{0.937 \pm 0.028}}$\\
Parkinson (195, 2) & $0.859 \pm 0.062$ & $0.848 \pm 0.064$ & $0.858 \pm 0.065$ & $\mathbf{\mathbf{0.863 \pm 0.065}}$\\
Planning Relax (182, 2) & $0.595 \pm 0.075$ & $0.725 \pm 0.049$ & $\mathbf{\mathbf{0.729 \pm 0.048}}$ & $0.595 \pm 0.075$\\
QSAR Biodegradation (1055, 2) & $0.752 \pm 0.031$ & $0.741 \pm 0.033$ & $0.757 \pm 0.026$ & $\mathbf{\mathbf{0.761 \pm 0.028}}$\\
Seeds (210, 3) & $0.918 \pm 0.034$ & $0.914 \pm 0.040$ & $0.905 \pm 0.081$ & $\mathbf{\mathbf{0.925 \pm 0.033}}$\\
Spambase (4601, 2) & $0.844 \pm 0.027$ & $0.839 \pm 0.028$ & $0.842 \pm 0.029$ & $\mathbf{\mathbf{0.846 \pm 0.026}}$\\
Vertebral Column 3C (310, 3) & $0.800 \pm 0.050$ & $0.725 \pm 0.139$ & $0.804 \pm 0.046$ & $\mathbf{\mathbf{0.819 \pm 0.044}}$\\
WFR24\textsuperscript{e} (5456, 4) & $\mathbf{\mathbf{0.995 \pm 0.002}}$ & $\mathbf{0.994 \pm 0.002}$ & $\mathbf{0.994 \pm 0.002}$ & $\mathbf{0.994 \pm 0.001}$\\
Wine (178, 3) & $\mathbf{\mathbf{0.908 \pm 0.041}}$ & $0.902 \pm 0.045$ & $0.903 \pm 0.043$ & $0.904 \pm 0.046$\\
Yeast (1484, 10) & $0.429 \pm 0.019$ & $0.368 \pm 0.059$ & $0.384 \pm 0.058$ & $\mathbf{\mathbf{0.442 \pm 0.019}}$\\
\bottomrule
\end{tabular}

\footnotesize \textsuperscript{a}Breast Cancer Wisconsin Diagnostic, \textsuperscript{b}Climate Model Simulation Crashes, \textsuperscript{c}Connectionist Bench Sonar,

\textsuperscript{d}Diabetic Retinopathy Debrecen, \textsuperscript{e}Wall Following Robot 24
\end{table}

Our algorithm performs better than or similarly to the other algorithms on 13 out of 19 datasets, and on 16 out of 19 when excluding the original unpruned tree.
Furthermore, our algorithm is able to do well on datasets of different sizes: it has the best performance on the Iris dataset with only 150 examples as well as on the Spambase dataset with 4601 examples.
The mean accuracy gain of our algorithm versus the CART algorithm is of 2.02\%, which suggests that it could be profitable to use our bound-based algorithm to prune trees instead of CART.
Another advantage of our pruning algorithm is that it is on average 19.5 times faster than the pruning process of CART, due to the fact that our algorithm does not rely on cross-validation.

While our algorithm works well in practice, it is unfortunate that the computed bound $\epsilon$ of the pruned tree is uninformative (\ie greater than 1) most of the time.
On the other hand, the good performances of our algorithm shows that Shawe-Taylor's bound~\eqref{eq:shawe-taylor_bound} and our bound~\eqref{eq:ub_partitioning_function_tree} capture the behavior of decision trees well, up to a possibly large constant factor.

It is interesting to see that our pruning algorithm and the CART algorithm do not perform the same trade-off; indeed, the final tree produced by CART has three times less leaves on average than the pruned tree generated by our algorithm.
This suggests that CART prunes decision trees more aggressively than necessary.

As for our modified version of CART, it generally does better than the original CART algorithm (it has a mean accuracy gain of 1.20\%), but it is not as good as the algorithm based on the bound, and as such is of limited interest.

\section{Conclusion}
By considering binary decision trees as partitioning machines, and introducing the set of partitioning functions of a tree class, we have found that the VC dimension of a tree class is given by the largest integer $d$ such that $\pi^2_T(d) = 2^{d-1}-1$.
Then, we found at tight upper bound on the 2-partitioning function of the class of decision stumps on $\ell$ real-valued features.
This bound allowed us to find the exact VC dimension of decision stumps, which is given by the largest $d$ such that $2\ell \geq \tbinom{d}{\floor{\frac{d}{2}}}$.
It was then possible to extend these results to yield a recursive upper bound of the $c$-partitioning functions of any class of binary decision tree.
As a corollary, we found that the VC dimenion of a tree class with $N$ internal nodes is of order $O(N\log(N\ell))$.
Based on our findings, we proposed a pruning algorithm which performed better or similarly to CART on 16 out of 19 datasets, showing that our bound-based algorithm is a viable alternative to CART.

In the future, we wish to extend our framework to decision trees on categorical features.
While our partitioning framework can also be applied to categorical features, there are some obstacles to overcome at first.
Most notably, as opposed to the case of real-valued features, there exist multiple ways to produce splitting rules on categorical features.
For example, ID3 \citep{quinlan1986induction} produces a subtree for each category, LightGBM \citep{ke2017lightgbm} bundles features together, and CART \citep{breiman1984classification} examines all possible split combinations.
Other techniques involve binary encodings such as one-versus-all or one-versus-one.
Every such way to proceed may result in different partitioning patterns requiring different analyses.
Furthermore, one must introduce new notation to be able to handle the specific feature distribution relevant to each problem, \textit{i.e.} there could be a certain number of features that are binary, another number that are ternary, and so on for all category sizes.
We think these difficulties can be resolved and we aim to do so in a subsequent paper.
% In the future, we wish to extend our framework to decision trees on categorical features and to derive a lower bound for the partitioning functions.
% Finally, it remains an open problem to know if the bound of Theorem~\ref{thm:ub_partitioning_func_decision_stumps} is in fact an equality for all $m$. 

\section*{Broader Impact}

This work could be profitable to machine learning practitioners that use decision trees to produce predictive models. 
%Our experiments show that in most cases, our pruning algorithm leads to better generalization than CART, one of the most commonly used pruning algorithms, and it is faster to run because it does not rely on cross-validation.
The methods and results presented in this work are not incompatible with methods that try to correct the bias present in some datasets and with machine learning fairness methods that should be applied when the learned model attempts to make predictions on some aspects of human behaviour.

% Authors are required to include a statement of the broader impact of their work, including its ethical aspects and future societal consequences. 
% Authors should discuss both positive and negative outcomes, if any. For instance, authors should discuss a) 
% who may benefit from this research, b) who may be put at disadvantage from this research, c) what are the consequences of failure of the system, and d) whether the task/method leverages
% biases in the data. If authors believe this is not applicable to them, authors can simply state this.

% Use unnumbered first level headings for this section, which should go at the end of the paper. {\bf Note that this section does not count towards the eight pages of content that are allowed.}

\begin{ack}
This work was supported in part by NSERC. We are grateful to Gaël Letarte for his comments and suggestions on preliminary versions.
\end{ack}

\bibliographystyle{plainnat}
\bibliography{neurips_2020}

\begin{thebibliography}{36}
\providecommand{\natexlab}[1]{#1}
\providecommand{\url}[1]{\texttt{#1}}
\expandafter\ifx\csname urlstyle\endcsname\relax
  \providecommand{\doi}[1]{doi: #1}\else
  \providecommand{\doi}{doi: \begingroup \urlstyle{rm}\Url}\fi

\bibitem[Aeberhard et~al.(1994)Aeberhard, Coomans, and
  De~Vel]{aeberhard1994comparative}
Stefan Aeberhard, Danny Coomans, and Olivier De~Vel.
\newblock Comparative analysis of statistical pattern recognition methods in
  high dimensional settings.
\newblock \emph{Pattern Recognition}, 27\penalty0 (8):\penalty0 1065--1077,
  1994.

\bibitem[Antal and Hajdu(2014)]{antal2014ensemble}
B{\'a}lint Antal and Andr{\'a}s Hajdu.
\newblock An ensemble-based system for automatic screening of diabetic
  retinopathy.
\newblock \emph{Knowledge-based systems}, 60:\penalty0 20--27, 2014.

\bibitem[Aslan et~al.(2009)Aslan, Yildiz, and Alpaydin]{aslan2009calculating}
Ozlem Aslan, Olcay~Taner Yildiz, and Ethem Alpaydin.
\newblock Calculating the {VC}-dimension of decision trees.
\newblock In \emph{2009 24th International Symposium on Computer and
  Information Sciences}, pages 193--198. IEEE, 2009.

\bibitem[Ayres-de Campos et~al.(2000)Ayres-de Campos, Bernardes, Garrido,
  Marques-de Sa, and Pereira-Leite]{ayres2000sisporto}
Diogo Ayres-de Campos, Joao Bernardes, Antonio Garrido, Joaquim Marques-de Sa,
  and Luis Pereira-Leite.
\newblock Sisporto 2.0: a program for automated analysis of cardiotocograms.
\newblock \emph{Journal of Maternal-Fetal Medicine}, 9\penalty0 (5):\penalty0
  311--318, 2000.

\bibitem[Berthonnaud et~al.(2005)Berthonnaud, Dimnet, Roussouly, and
  Labelle]{berthonnaud2005analysis}
Eric Berthonnaud, Joann{\`e}s Dimnet, Pierre Roussouly, and Hubert Labelle.
\newblock Analysis of the sagittal balance of the spine and pelvis using shape
  and orientation parameters.
\newblock \emph{Clinical Spine Surgery}, 18\penalty0 (1):\penalty0 40--47,
  2005.

\bibitem[Bhatt(2012)]{bhatt2012planning}
Rajen Bhatt.
\newblock Planning-relax dataset for automatic classification of eeg signals.
\newblock \emph{UCI Machine Learning Repository}, 2012.

\bibitem[B{\'o}na(2015)]{bona2015handbook}
Mikl{\'o}s B{\'o}na.
\newblock \emph{Handbook of enumerative combinatorics}, volume~87.
\newblock CRC Press, 2015.

\bibitem[Breiman et~al.(1984)Breiman, Friedman, Stone, and
  Olshen]{breiman1984classification}
Leo Breiman, Jerome Friedman, Charles~J Stone, and Richard~A Olshen.
\newblock \emph{Classification and regression trees}.
\newblock CRC press, 1984.

\bibitem[Charytanowicz et~al.(2010)Charytanowicz, Niewczas, Kulczycki,
  Kowalski, {\L}ukasik, and {\.Z}ak]{charytanowicz2010complete}
Ma{\l}gorzata Charytanowicz, Jerzy Niewczas, Piotr Kulczycki, Piotr~A Kowalski,
  Szymon {\L}ukasik, and S{\l}awomir {\.Z}ak.
\newblock Complete gradient clustering algorithm for features analysis of x-ray
  images.
\newblock In \emph{Information technologies in biomedicine}, pages 15--24.
  Springer, 2010.

\bibitem[Corless et~al.(1996)Corless, Gonnet, Hare, Jeffrey, and
  Knuth]{corless96lambertw}
R.~M. Corless, G.~H. Gonnet, D.~E.~G. Hare, D.~J. Jeffrey, and D.~E. Knuth.
\newblock On the {Lambert} {$W$} function.
\newblock \emph{Advances in Computational Mathematics}, 5\penalty0
  (1):\penalty0 329--359, Dec 1996.
\newblock ISSN 1572-9044.
\newblock \doi{10.1007/BF02124750}.

\bibitem[Drouin et~al.(2019)Drouin, Letarte, Raymond, Marchand, Corbeil, and
  Laviolette]{drouin2019interpretable}
Alexandre Drouin, Ga{\"e}l Letarte, Fr{\'e}d{\'e}ric Raymond, Mario Marchand,
  Jacques Corbeil, and Fran{\c{c}}ois Laviolette.
\newblock Interpretable genotype-to-phenotype classifiers with performance
  guarantees.
\newblock \emph{Scientific reports}, 9\penalty0 (1):\penalty0 1--13, 2019.

\bibitem[Dua and Graff(2017)]{Dua:2019}
Dheeru Dua and Casey Graff.
\newblock {UCI} machine learning repository, 2017.
\newblock URL \url{http://archive.ics.uci.edu/ml}.

\bibitem[Fisher(1936)]{fisher1936use}
Ronald~A Fisher.
\newblock The use of multiple measurements in taxonomic problems.
\newblock \emph{Annals of eugenics}, 7\penalty0 (2):\penalty0 179--188, 1936.

\bibitem[Freire et~al.(2009)Freire, Barreto, Veloso, and
  Varela]{freire2009short}
Ananda~L Freire, Guilherme~A Barreto, Marcus Veloso, and Antonio~T Varela.
\newblock Short-term memory mechanisms in neural network learning of robot
  navigation tasks: A case study.
\newblock In \emph{2009 6th Latin American Robotics Symposium (LARS 2009)},
  pages 1--6. IEEE, 2009.

\bibitem[Gey(2018)]{gey2018vapnik}
Servane Gey.
\newblock Vapnik--{Chervonenkis} dimension of axis-parallel cuts.
\newblock \emph{Communications in Statistics-Theory and Methods}, 47\penalty0
  (9):\penalty0 2291--2296, 2018.

\bibitem[Gil et~al.(2012)Gil, Girela, De~Juan, Gomez-Torres, and
  Johnsson]{gil2012predicting}
David Gil, Jose~Luis Girela, Joaquin De~Juan, M~Jose Gomez-Torres, and Magnus
  Johnsson.
\newblock Predicting seminal quality with artificial intelligence methods.
\newblock \emph{Expert Systems with Applications}, 39\penalty0 (16):\penalty0
  12564--12573, 2012.

\bibitem[Gorman and Sejnowski(1988)]{gorman1988analysis}
R~Paul Gorman and Terrence~J Sejnowski.
\newblock Analysis of hidden units in a layered network trained to classify
  sonar targets.
\newblock \emph{Neural networks}, 1\penalty0 (1):\penalty0 75--89, 1988.

\bibitem[Graham et~al.(1989)Graham, Knuth, Patashnik, and
  Liu]{graham1989concrete}
Ronald~L. Graham, Donald~E. Knuth, Oren Patashnik, and Stanley Liu.
\newblock Concrete mathematics: a foundation for computer science.
\newblock \emph{Computers in Physics}, 3\penalty0 (5):\penalty0 106--107, 1989.

\bibitem[Haberman(1976)]{haberman1976generalized}
Shelby~J Haberman.
\newblock Generalized residuals for log-linear models.
\newblock In \emph{Proceedings of the 9th international biometrics conference},
  pages 104--122, 1976.

\bibitem[Hall(1935)]{HallMarriage}
P.~Hall.
\newblock On representatives of subsets.
\newblock \emph{Journal of the London Mathematical Society}, s1-10\penalty0
  (1):\penalty0 26--30, 1935.
\newblock \doi{10.1112/jlms/s1-10.37.26}.
\newblock URL
  \url{https://londmathsoc.onlinelibrary.wiley.com/doi/abs/10.1112/jlms/s1-10.37.26}.

\bibitem[Horton and Nakai(1996)]{horton1996probabilistic}
Paul Horton and Kenta Nakai.
\newblock A probabilistic classification system for predicting the cellular
  localization sites of proteins.
\newblock In \emph{Ismb}, volume~4, pages 109--115, 1996.

\bibitem[Ke et~al.(2017)Ke, Meng, Finley, Wang, Chen, Ma, Ye, and
  Liu]{ke2017lightgbm}
Guolin Ke, Qi~Meng, Thomas Finley, Taifeng Wang, Wei Chen, Weidong Ma, Qiwei
  Ye, and Tie-Yan Liu.
\newblock Lightgbm: A highly efficient gradient boosting decision tree.
\newblock In \emph{Advances in neural information processing systems}, pages
  3146--3154, 2017.

\bibitem[Little et~al.(2007)Little, McSharry, Roberts, Costello, and
  Moroz]{little2007exploiting}
Max~A Little, Patrick~E McSharry, Stephen~J Roberts, Declan~AE Costello, and
  Irene~M Moroz.
\newblock Exploiting nonlinear recurrence and fractal scaling properties for
  voice disorder detection.
\newblock \emph{Biomedical engineering online}, 6\penalty0 (1):\penalty0 23,
  2007.

\bibitem[Lucas et~al.(2013)Lucas, Klein, Tannahill, Ivanova, Brandon,
  Domyancic, and Zhang]{lucas2013failure}
DD~Lucas, R~Klein, J~Tannahill, D~Ivanova, S~Brandon, D~Domyancic, and Y~Zhang.
\newblock Failure analysis of parameter-induced simulation crashes in climate
  models.
\newblock \emph{Geoscientific Model Development}, 6\penalty0 (4):\penalty0
  1157--1171, 2013.

\bibitem[Maimon and Rokach(2002)]{maimon2002improving}
Oded Maimon and Lior Rokach.
\newblock Improving supervised learning by feature decomposition.
\newblock In \emph{International Symposium on Foundations of Information and
  Knowledge Systems}, pages 178--196. Springer, 2002.

\bibitem[Mansour(1997)]{mansour1997pessimistic}
Yishay Mansour.
\newblock Pessimistic decision tree pruning based on tree size.
\newblock In \emph{Proceedings of the Fourteenth International Conference on
  Machine Learning}, pages 195--201. Morgan Kaufmann, 1997.

\bibitem[Mansouri et~al.(2013)Mansouri, Ringsted, Ballabio, Todeschini, and
  Consonni]{mansouri2013quantitative}
Kamel Mansouri, Tine Ringsted, Davide Ballabio, Roberto Todeschini, and Viviana
  Consonni.
\newblock Quantitative structure--activity relationship models for ready
  biodegradability of chemicals.
\newblock \emph{Journal of chemical information and modeling}, 53\penalty0
  (4):\penalty0 867--878, 2013.

\bibitem[Marchand and Sokolova(2005)]{ms-05}
Mario Marchand and Marina Sokolova.
\newblock Learning with decision lists of data-dependent features.
\newblock \emph{Journal of Machine Learning Research}, 6\penalty0
  (Apr):\penalty0 427--451, 2005.

\bibitem[M{\"u}tze et~al.(2018)M{\"u}tze, Nummenpalo, and
  Walczak]{mutze2018sparse}
Torsten M{\"u}tze, Jerri Nummenpalo, and Bartosz Walczak.
\newblock Sparse kneser graphs are hamiltonian.
\newblock In \emph{Proceedings of the 50th Annual ACM SIGACT Symposium on
  Theory of Computing}, pages 912--919. ACM, 2018.

\bibitem[Quinlan(1986)]{quinlan1986induction}
J.~Ross Quinlan.
\newblock Induction of decision trees.
\newblock \emph{Machine learning}, 1\penalty0 (1):\penalty0 81--106, 1986.

\bibitem[Shawe-Taylor et~al.(1998)Shawe-Taylor, Bartlett, Williamson, and
  Anthony]{shawe1998structural}
John Shawe-Taylor, Peter~L. Bartlett, Robert~C. Williamson, and Martin Anthony.
\newblock Structural risk minimization over data-dependent hierarchies.
\newblock \emph{IEEE transactions on Information Theory}, 44\penalty0
  (5):\penalty0 1926--1940, 1998.

\bibitem[Sigillito et~al.(1989)Sigillito, Wing, Hutton, and
  Baker]{sigillito1989classification}
Vincent~G Sigillito, Simon~P Wing, Larrie~V Hutton, and Kile~B Baker.
\newblock Classification of radar returns from the ionosphere using neural
  networks.
\newblock \emph{Johns Hopkins APL Technical Digest}, 10\penalty0 (3):\penalty0
  262--266, 1989.

\bibitem[Simon(1991)]{simon1991vapnik}
Hans~Ulrich Simon.
\newblock The {V}apnik-{C}hervonenkis dimension of decision trees with bounded
  rank.
\newblock \emph{Information Processing Letters}, 39\penalty0 (3):\penalty0
  137--141, 1991.

\bibitem[Street et~al.(1993)Street, Wolberg, and
  Mangasarian]{street1993nuclear}
W~Nick Street, William~H Wolberg, and Olvi~L Mangasarian.
\newblock Nuclear feature extraction for breast tumor diagnosis.
\newblock In \emph{Biomedical image processing and biomedical visualization},
  volume 1905, pages 861--870. International Society for Optics and Photonics,
  1993.

\bibitem[Vapnik(1998)]{vapnik1998statistical}
Vladimir Vapnik.
\newblock \emph{{S}tatistical {L}earning {T}heory}.
\newblock John {W}iley \& {S}ons, 1998.

\bibitem[Y{\i}ld{\i}z(2015)]{yildiz2015vc}
Olcay~Taner Y{\i}ld{\i}z.
\newblock {VC}-dimension of univariate decision trees.
\newblock \emph{IEEE transactions on neural networks and learning systems},
  26\penalty0 (2):\penalty0 378--387, 2015.

\end{thebibliography}

\clearpage
\appendix

\section{Proof of Theorem~\ref{thm:ub_partitioning_func_decision_stumps}}
\label{app:proof_theorem_decision_stump}

Before proceeding with the proof, we introduce a convenient way to think about a node's decision rule. Recall that a node is associated with a rule described by a feature $i \in [\ell]$, a threshold $\theta \in \mathds{R}$, and a sign $s \in \cb{\pm 1}$. The sample $S = \{\x_1, \hdots, \x_m\}$ may be represented by a collection $\Sigma$ of $\ell$ permutations of $[m]$ representing the ordering of its data points according to their values for each feature, since that relative ordering encapsulates all pertinent information on the sample, from the perspective of decision trees. To be more precise, for each $i = 1, \hdots, \ell$, let $\sigma^{i}$ be a permutation of $[m]$ satisfying 
\[
x_{\sigma_1^{i}}^{i} \leq x_{\sigma_2^{i}}^{i} \leq \cdots \leq x_{\sigma_m^{i}}^{i} \, .
\]
In general, unless the data points all have different values for a given feature, there may be many such permutations; just pick one arbitrarily. 

Any node in a decision tree splits the data points in two according to a rule of the form 
\[
t(\x) = \begin{cases} t_l(\x) &\text{if} \, \sign(x^{i} - \theta) = s \\
t_r(\x) &\text{otherwise.}
\end{cases}
\]
This corresponds to splitting the permutation 
\[
\sigma^{i} = \begin{bmatrix}
\sigma_1^{i} & \sigma_2^{i} & \cdots & \sigma_m^{i}
\end{bmatrix}
\]
in two parts, sending examples $\x_{\sigma_j^i}$ to one subtree for $j \leq J$, and sending the rest of the examples to the other subtree, where $J$ is determined by $\theta$ and $s$. In fact, as long as the inequalities 
\[
x_{\sigma_1^{i}}^{i} < x_{\sigma_2^{i}}^{i} < \cdots < x_{\sigma_m^{i}}^{i} 
\]
are strict (all data points have different values for each feature), then all the different ways of splitting the (now unique) permutation $\sigma^{i}$ induce a split on the sample $S$ according to which it was defined. This situation could be called the worst-case scenario, because it allows for more distinct 2-partitions to be realized on the sample. 

We split the proof in 4 parts: 1) the bound itself, 2) the equality for $2\ell \leq m$, 3) the equality for $2\ell \geq \binom{m}{\floor{\frac{m}{2}}}$, and 4) the equality for $1 \leq m \leq 7$.

% \subsection{Proof of Lemma~\ref{lem:partitions_node}}
% \label{app:partitions_node_proof}
\subsection{Proof of part 1 of Theorem~\ref{thm:ub_partitioning_func_decision_stumps}}
\label{app:partitions_node}

We want to show that 
\[
\pi^2_T(m) \leq \frac{1}{2} \sum_{k=1}^{m-1} \min \cb{ 2\ell, \binom{m}{k} }.
\]
where $T$ is the class of decision stumps on $\ell$ real-valued features.

\begin{proof}
First, let $\R(S)$ be the set of 2-partitions of $S$ realizable by a single node, and notice that bounding the cardinality of $\R(S)$ directly gives a bound on $\pi^2_T(m)$ if the bound does not depend directly on $S$.
%Thus, we want to bound $\abs{\R(S)}$ as a function of $m$ only.

Let $\R_k(S) \subset \R(S)$ be the subset of $2$-partitions with a part of size $k$, and notice $\R_k(S) = \R_{m-k}(S)$.
Therefore, we can decompose $\R(S)$ into the disjoint union
\begin{align}\label{eq:root_decomposition}
    \R(S) = \bigcup_{k=1}^{\floor{\frac{m}{2}}} \R_k(S).
\end{align}

To bound $\abs{\R_k(S)}$, first consider $k < \frac{m}{2}$. Every partition in $\R_k(S)$ is determined by a set of $k$ data points, so that $\abs{\R_k(S)} \leq \binom{m}{k}$, the number of $k$-subsets of $S$. 
On the other hand, given a feature $i \in [\ell]$, in the worst-case scenario, we can split the permutation $\sigma^{i}$ after the $k$ first points or before the $k$ last points to induce 2 distinct elements of $\R_k(S)$. Since there are $\ell$ features, this makes a total of at most $2\ell$ realizable 2-partitions with a part of size $k$. We conclude that, for $k < \frac{m}{2}$, we have $\abs{\R_k(S)} \leq \min \left\{ 2\ell, \binom{m}{k} \right\}$. 

Now let $k = \frac{m}{2}$. Then the same arguments apply, except that the number of $2$-partitions with a part of size $k$ is $\frac{1}{2} \binom{m}{k}$ because each such partition contains two subsets of the same size $k$. Moreover, for the same reason, the node can produce at most only one $2$-partition with a part of size $k$ for each feature. 
Thus, $\abs{\R_k(S)} \leq \min \left\{ \ell, \frac{1}{2} \binom{m}{k} \right\}$.

Combining our results, we have
\begin{align}\label{eq:bound_on_RkS}
    \abs{\R_k(S)} \leq \begin{cases}
        \min \cb{\ell, \frac{1}{2}\binom{m}{k} } & \text{if} \, k = \frac{m}{2} \\
        \min \cb{2\ell, \binom{m}{k} } & \text{otherwise.}
    \end{cases}
\end{align}

Using Inequality~\eqref{eq:bound_on_RkS}, the symmetry $\R_k(S) = \R_{m-k}(S)$ yields
\[
\abs{\R(S)} = \sum_{k=1}^{\floor{\frac{m}{2}}} \abs{\R_k(S)} \leq \sum_{k = 1}^{m - 1} \min \left\{ 2\ell, \binom{m}{k} \right\} 
\]
which concludes the proof, since the bound on $\abs{\R(S)}$ depends only on $m$ and not on $S$.
\end{proof}

\subsection{Proof of part 2 of Theorem~\ref{thm:ub_partitioning_func_decision_stumps}}
\label{app:proof_part_2_vcdim_stump}

\begin{proof}
We want to show that the bound of Theorem~\ref{thm:ub_partitioning_func_decision_stumps} is an equality for $2\ell \leq m$. To this end, we want to show the existence of a sample $S$ such that 
\[
\lvert \mathcal{R}_k(S) \rvert = \begin{cases} \ell &\text{if} \, k = \frac{m}{2} \\
2\ell &\text{otherwise.}
\end{cases}
\]
Since $2\ell \leq m$ implies $2\ell \leq \binom{m}{k}$ for all $k$, we will have
\[
\lvert \mathcal{R}(S) \rvert = \sum_{k = 1}^{\floor{\frac{m}{2}}} \lvert \mathcal{R}_k(S) \rvert = \ell (m - 1) = \frac{1}{2} \sum_{k = 1}^{m - 1} 2\ell = \frac{1}{2} \sum_{k = 1}^{m - 1} \min\left\{2\ell, \binom{m}{k}\right\}
\]
which establishes that the bound of Theorem~\ref{thm:ub_partitioning_func_decision_stumps} is an equality. 

Let us construct a suitable sample $S$. Consider the permutations $\sigma^1, \hdots, \sigma^\ell$ given by the rows of the following permutation representation of $S$:
\[
\Sigma =
\left[
\begin{array}{c c c c | c c c c | c c c c}
1 & 2 & \hdots & l & 2l + 1 & 2l + 2 & \hdots & m & 2l & 2l - 1 & \hdots & l + 1\\
2 & 3 & \hdots & l + 1 & 2l + 1 & 2l + 2 & \hdots & m & 1 & 2l & \hdots & l + 2\\
3 & 4 & \hdots & l + 2 & 2l + 1 & 2l + 2 & \hdots & m & 2 & 1 & \hdots & l + 3\\
\vdots & \vdots & \ddots & \vdots & \vdots & \vdots & \ddots & \vdots & \vdots & \vdots & \ddots & \vdots\\
l & l + 1 & \hdots & 2l - 1 & 2l + 1 & 2l + 2 & \hdots & m & l - 1 & l - 2 & \hdots & 2l
\end{array} 
\right] .
\]
$\Sigma$ is built up from an $\ell \times \ell$ matrix on the left, an $\ell \times (m - 2\ell)$ matrix in the middle, and an $\ell \times \ell$ matrix on the right. In the remainder of this paragraph, a shift is a shift in the sequence $1, 2, \hdots, 2\ell$. The first row of the left matrix is $1, 2, \hdots, \ell$; subsequent rows are obtained by shifting one position to the right. The middle matrix has identical rows running from $2\ell + 1$ to $m$. The first row of the right matrix is $2\ell, 2\ell - 1, \hdots, \ell + 1$; subsequent rows are obtained by shifting one position to the left. For example, if $\ell = 3$ and $m = 9$, we have
\[ 
\Sigma =
\left[
\begin{array}{c c c | c c c | c c c}
1 & 2 & 3 & 7 & 8 & 9 & 6 & 5 & 4\\
2 & 3 & 4 & 7 & 8 & 9 & 1 & 6 & 5\\
3 & 4 & 5 & 7 & 8 & 9 & 2 & 1 & 6
\end{array} 
\right] .
\]
It is clear that, for $k = 1, \hdots, \floor{\frac{m}{2}}$, splitting any of these permutations after the first $k$ points or before the last $k$ points always induces different 2-partitions with a part of size $k$ on the sample, as long as the sample is chosen so that the strict inequalities 
\[
x_{\sigma_1^{i}}^{i} < x_{\sigma_2^{i}}^{i} < \cdots < x_{\sigma_m^{i}}^{i} 
\]
hold; it suffices to choose $x_{\sigma_j^{i}}^{i} = j$ for $i = 1, \hdots, \ell$ and $j = 1, \hdots, m$. This gives us a total of $\ell$ distinct 2-partitions if $k = \frac{m}{2}$ (with even $m$), and a total of $2\ell$ distinct permutations if $k < \frac{m}{2}$, as required. 
\end{proof}

\subsection{Proof of part 3 of Theorem~\ref{thm:ub_partitioning_func_decision_stumps}}
\label{app:proof_part_3_vcdim_stump}

We prove part 3 of Theorem~\ref{thm:ub_partitioning_func_decision_stumps} by showing that for $2\ell \geq \tbinom{m}{\floormtwo}$ \big(so that $2\ell \geq \binom{m}{k}$ for all $k$\big), there exists a sample $S$ such that 
\[
\abs{\R(S)} = \sum_{k = 1}^{m - 1} \binom{m}{k} = \sum_{k = 1}^{m - 1} \min \left\{ 2\ell, \binom{m}{k} \right\}.
\]

We proceed in two steps. First, we show that there exists a sample $S$ of $m$ examples on which every 2-partition with a part of size $\floormtwo$ is realized by a stump, when $2\ell \geq \binom{m}{\floormtwo}$. 
Second, we use induction from this base case to establish the proof for all part sizes.
More precisely, we show that if there exists a sample $S_k$ such that a stump can realize every $2$-partition with a part of size $2 \leq k \leq \frac{m}{2}$, then there also exists a sample $S_{k - 1}$ of the same size such that a stump can realize every $2$-partition with a part of size $k$ \emph{and} every $2$-partition with a part of size $k-1$.

Let $\Sigma$ be the permutation representation of $S$, as explained at the beginning of Appendix~\ref{app:proof_theorem_decision_stump}. Furthermore, assume we are in the worst-case scenario where
\[
x_{\sigma_1^{i}}^{i} < x_{\sigma_2^{i}}^{i} < \cdots < x_{\sigma_m^{i}}^{i} 
\]
for all $i \in [\ell]$. In this case, showing that every 2-partition of $S$ is realizable by a decision stump is equivalent to showing that every $k$-subset of $[m]$ is attainable by splitting a permutation of $\Sigma$ in two, either by splitting after the first $k$ elements or before the last $k$ elements for every possible $k$. Moreover, we only need to consider $k$-subsets for $1 \leq k \leq \frac{m}{2}$ since $\R_k(S) = \R_{m-k}(S)$.

\textbf{Step 1.}
We want to show that there exists a sample $S_\floormtwo$ of $m$ examples on which every $2$-partition with a part of size $\floormtwo$ is realized by a stump when $2\ell \geq \tbinom{m}{\floormtwo}$, \ie when $\ell \geq \ceil{\frac{1}{2}\tbinom{m}{\floormtwo}}$. Let $\Sigma_{\floor{\frac{m}{2}}}$ be its permutation representation. Our problem is then equivalent to finding a matrix $\Sigma_{\floor{\frac{m}{2}}}$ whose rows are permutations of $[m]$ such that each $\floor{\frac{m}{2}}$-subset of $[m]$ may be found as the first $\floor{\frac{m}{2}}$ elements or the last $\floor{\frac{m}{2}}$ elements of a row of $\Sigma_{\floor{\frac{m}{2}}}$. 

This is easy for even $m$. Given that $\ell \geq \frac{1}{2}\tbinom{m}{\frac{m}{2}}$ and that there are exactly $\frac{1}{2}\binom{m}{\frac{m}{2}}$ different 2-partitions of $[m]$ with a part of size $\frac{m}{2}$, we can fit them all the first $\ell$ rows of the matrix $\Sigma_{\frac{m}{2}}$ with the first $\frac{m}{2}$ elements of each row being the elements of the first part of each 2-partition. Then, $\Sigma_{\frac{m}{2}}$ induces a sample $S_{\frac{m}{2}}$ on which every 2-partition is realizable by a stump. If $S_{\frac{m}{2}} = \{ \x_1, \hdots, \x_m \}$, choosing $x_{\rho_j^{i}}^{i} = j$, where the $\rho_j^{i}$ are 
the elements of the matrix $\Sigma_{\frac{m}{2}}$, suffices. 

Now, let's see what happens when $m$ is odd. Consider the minimal case $\ell = \ceil{\frac{1}{2}\tbinom{m}{\floormtwo}}$. We rephrase our problem as a graph problem. 
Let the vertices of the graph $G = (V, E)$ be the $\floormtwo$-subsets of $[m]$ and only place edges between disjoint $\floor{\frac{m}{2}}$-subsets. Now, pairs of $\floor{\frac{m}{2}}$-subsets with an edge connecting them are exactly the pairs of $\floor{\frac{m}{2}}$-subsets of $[m]$ whose elements can occur in the same row of $\Sigma_{\floor{\frac{m}{2}}}$ (since each row is a permutation and therefore contains each element of $[m]$ exactly once). 
The problem of constructing a suitable matrix $\Sigma_{\floor{\frac{m}{2}}}$ becomes equivalent to showing that there exists a subset of edges $M\subseteq E$ such that no two edges $e_1, e_2 \in M$ are incident to the same vertex, with cardinality $\abs{M} = \ell$ if $\binom{m}{\floormtwo}$ is even and $\abs{M} = \ell - 1$ if $\binom{m}{\floormtwo}$ is odd (since in this case, one $\floor{\frac{m}{2}}$-subset of $[m]$ will have its own row in the matrix $\Sigma_{\floor{\frac{m}{2}}}$). Such problems are called \emph{matching} problems in the field of graph theory.

\begin{figure}
\centering
\begin{tikzpicture}
\def\outerradius{2.25}
\def\innerradius{1}
\foreach \angle/\tag in {0/12, 72/45, 144/23, 216/15, 288/34}{
    \node[circle, draw, inner sep=2pt](\tag) at (90+\angle:\outerradius) {\footnotesize\tag};
}
\foreach \angle/\tag in {0/35, 72/13, 144/14, 216/24, 288/25}{
    \node[circle, draw, inner sep=2pt](\tag) at (90+\angle:\innerradius) {\footnotesize\tag};
}
\draw (12) -- (34);
\draw (12) -- (45);
\draw[ultra thick, color3] (12) -- (35);

\draw[ultra thick, color3] (34) -- (25);
\draw (34) -- (15);

\draw (35) -- (24);
\draw (35) -- (14);

\draw[ultra thick, color3] (45) -- (13);
\draw (45) -- (23);

\draw (25) -- (13);
\draw (25) -- (14);

\draw[ultra thick, color3] (15) -- (24);
\draw (15) -- (23);

\draw (24) -- (13);

\draw[ultra thick, color3] (14) -- (23);

\end{tikzpicture}
\caption{The odd graph $O_2$, also commonly known as the Petersen Graph. One matching of size $5$ is shown in bold red.}
\label{fig:petersen_graph}
\end{figure}
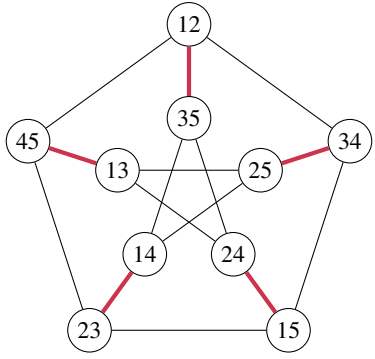

As it turns out, the graph $G$ is known as the \textit{Odd Graph} $O_n$ with $n=\floor{\frac{m}{2}}$ (since $m = 2\floor{\frac{m}{2}} + 1$ when $m$ is odd).
According to \cite{mutze2018sparse}, $O_n$ has at least one Hamiltonian cycle for $n=1$ and for every $n\geq 3$, a Hamiltonian cycle being a cycle which goes through every vertex exactly once. In particular, it has a Hamiltonian path as long as $n \neq 2$. 
This implies that for $n \neq 2$, there exists a matching of size $\floor{ \frac{1}{2} \binom{m}{ \floor{\frac{m}{2}} } }$.
Indeed, it suffices to take one such Hamiltonian path, add the first edge to $M$, skip the next one, and continue adding every other edge to $M$ as we follow along the path. This ensures that every vertex is incident to exactly one of the selected edges, except when the number of vertices is odd, in which case one vertex is left out (thus accounting for the floor function).
The case $n=2$ (which only occurs when $m = 5$) is exceptional and $O_2$ corresponds to the Petersen Graph, which has no Hamiltonian cycle.
However, from Figure~\ref{fig:petersen_graph}, we can see that there still exists a matching of size $\ell=\frac{1}{2} \binom{5}{ \floor{\frac{5}{2}} }=5$.

We can easily construct a set of $\ell$ permutations which separates all $\floor{\frac{m}{2}}$-subsets from such a matching.
Pair off the $\floor{\frac{m}{2}}$-subsets which are joined by an edge in the chosen matching $M$. 
Since $m$ is odd, choosing a pair of disjoint $\floor{\frac{m}{2}}$-subsets of $[m]$ fixes $m-1$ elements of a permutation, leaving exactly one possible element to complete it.
Hence, sandwich the missing elements between each pair of $\floor{\frac{m}{2}}$-subsets to construct the rows of $\Sigma_{\floor{\frac{m}{2}}}$. 
If $\binom{m}{ \floor{\frac{m}{2}} }$ is even, we are done. 
Otherwise, put the last $\floor{\frac{m}{2}}$-subset at the beginning of the $\ell$-th row of $\Sigma_{\floor{\frac{m}{2}}}$. 
Lastly, if $\ell > \ceil{\frac{1}{2}\tbinom{m}{\frac{m}{2}}}$, then build the first $\ell$ rows of $\Sigma_{\floor{\frac{m}{2}}}$ as described above and fill in the rest with arbitrary permutations.
With this configuration, just like in the even case, $\Sigma_{\floor{\frac{m}{2}}}$ induces a sample $S_{\frac{m}{2}}$ on which every 2-partition is realizable by a stump. 

\textbf{Step 2.}
We want to prove that given a sample $S_\floormtwo$ on which every $2$-partition with a part of size $\floormtwo$ is realizable by a stump, we can construct a sample $S_1$ on which every $2$-partition is realizable by a stump.
We proceed inductively, showing that given a sample $S_k$ with $1 < k \leq \floor{\frac{m}{2}}$ on which every $2$-partition with a part of size $k, k+1, \dots, \floormtwo$ is realizable by a stump, there exists a sample $S_{k-1}$ of the same size as $S_k$ on which every $2$-partition with a part of size $k-1, k, k+1, \dots, \floormtwo$ is realizable by a stump.

Let $S_k$ be a sample such that no two of its instances have the same value for any feature, and let $\Sigma_k$ be its permutation representation. 

Then, Lemma~\ref{lem:injective_mapping_from_Ak_to_Ak+1}, proved below, assures us that there exists an injective map $\phi$ from the set $\binom{[m]}{k-1}$ of all $(k-1)$-subsets of $[m]$ to the set $\binom{[m]}{k}$ of all $k$-subsets of $[m]$ such that for every $(k-1)$-subset $a$, we have $a \subset \phi(a)$.

By assumption, $\Sigma_k$ separates all $k$-subsets of $[m]$, that is all $k$-subsets of $[m]$ appear either as the first $k$ elements or the last $k$ elements of a row of $\Sigma_k$. For some $a \in \binom{[m]}{k-1}$, reorder the elements of $\phi(a)$ appearing at the beginning or the end of a row of $\Sigma_k$ so that the elements of $a$ are either at the beginning or at the end of this row (according to whether the elements of $\phi(a)$ are at the beginning or at the end of the row). Notice that after this procedure, the new matrix $\Sigma'_k$ that is obtained still separates $\phi(a)$; moreover, it also separates $a$. Since the map $\phi$ is injective, we can continue this process without ever needing to reorder the same half-row twice, applying the same steps for each $a \in \binom{[m]}{k-1}$. This yields a final matrix $\Sigma_{k - 1}$ which induces the desired sample $S_{k - 1}$. 

Now, since Lemma~\ref{lem:injective_mapping_from_Ak_to_Ak+1} is valid for $2 \leq k \leq \floor{\frac{m}{2}}$, and because $S_\floormtwo$ is a set on which every $2$-partition with a part of size $\floormtwo$ can be realized by a stump, one can repeat the process above until $k=2$ so that $\R(S_1)$ contains every 2-partition.
Thus, $S_1$ is the set needed to conclude the proof.

\begin{lemma}
\label{lem:injective_mapping_from_Ak_to_Ak+1}
Let $\binom{[m]}{k} \eqdef \left\{ a \subseteq [m] : |a| = k \right\}$ be the set of all $k$-subsets of $[m]$.
Then, for $1 \leq k < \frac{m}{2}$ there exists an injective mapping $\phi : \binom{[m]}{k} \to \binom{[m]}{k+1}$ such that $a \subset \phi(a)$ for all $a \in \binom{[m]}{k}$.
\end{lemma}
\begin{proof}
Let $k$ be such that $1 \leq k < \frac{m}{2}$. Consider the bipartite graph $G = (V, E)$ whose set of vertices is $V = \binom{[m]}{k} \cup \binom{[m]}{k + 1}$, with an edge connecting $a \in \binom{[m]}{k}$ and $b \in \binom{[m]}{k + 1}$ if and only if $a \subset b$, and no other edges. The lemma is equivalent to finding a matching of $G$ which covers $\binom{[m]}{k}$ in the sense that each vertex in $\binom{[m]}{k}$ is incident to an edge of the matching. We show the existence of such a matching using Hall's marriage theorem (see \cite{HallMarriage}). 

Let $W \subseteq \binom{[m]}{k}$ and consider the set $N(W)$ containing all the vertices in $\binom{[m]}{k + 1}$ which are adjacent to a vertex in $W$, that is all $(k + 1)$-subsets of $[m]$ which contain a $k$-subset of $[m]$ from $W$. 

Given $a \in W$, we can make $m - k$ different $(k + 1)$-subsets containing $a$ by adding one of the $m - k$ elements of $[m]$ not present in $a$ to it. Since we can do this for each $a \in W$, we obtain $(m - k)\abs{W}$ (not necessarily all distinct) $(k + 1)$-subsets. In fact, in the worst case, when all $\binom{k + 1}{k} = k + 1$ different $k$-subsets of some $b \in \binom{[m]}{k + 1}$ are present in $W$, $b$ will be counted $k + 1$ times. Therefore $(k + 1)\abs{N(W)} \geq (m - k)\abs{W}$. Moreover, since $1 \leq k < \frac{m}{2}$, we have $m - k \geq k + 1$. This means
\[
\abs{N(W)} \geq \frac{m - k}{k + 1}\abs{W} \geq \abs{W}.
\]
Since this inequality holds for all $W \subseteq \binom{[m]}{k}$, a straightforward application of Hall's marriage theorem yields a matching of $G$ which covers $\binom{[m]}{k}$ and proves the lemma. 
\end{proof}

\subsection{Proof of part 4 of Theorem~\ref{thm:ub_partitioning_func_decision_stumps}}
\label{app:proof_m=1to7}

We now prove that
\begin{align}\label{eq:decision_stump_partitioning_function_app}
    \pi^2_T(m) = \frac{1}{2} \sum_{k=1}^{m-1} \min \cb{ 2\ell, \binom{m}{k} }
\end{align}
when $1 \leq m \leq 7$.
To do so, consider the permutation representation $\Sigma$ of a sample $S$ as described at the beginning of the Appendix.
We explicitly define $\Sigma$ which induces a sample $S$ that shows Equation~\eqref{eq:decision_stump_partitioning_function_app} is satisfied. 

One must understand the following matrices as follows.
If $\ell$ is less than or equal to the total number of rows of the matrix, build $\Sigma$ from the first $\ell$ rows.
If $\ell$ is greater than the number of rows of the matrix, add arbitrary permutations to fill out the rest of the rows of $\Sigma$; these do not matter because $\Sigma$ already separates all subsets of $[m]$ with its first $\ell$ rows.

\begin{itemize}
  \item $m=1$:
\begin{align*}
    \begin{bmatrix}
        1
    \end{bmatrix}
\end{align*}

\item $m=2$:
\begin{align*}
    \begin{bmatrix}
        1 & 2
    \end{bmatrix}
\end{align*}

  \item $m=3$:

\begin{align*}
    \begin{bmatrix}
        1 & 2 & 3\\
        1 & 3 & 2
    \end{bmatrix}
\end{align*}

  \item $m=4$:

\begin{align*}
    \begin{bmatrix}
        1 & 2 & 4 & 3\\
        2 & 3 & 1 & 4\\
        1 & 3 & 2 & 4
    \end{bmatrix}
\end{align*}

  \item $m=5$:

\begin{align*}
\begin{bmatrix}
1 & 2 & 3 & 5 & 4\\
2 & 3 & 4 & 1 & 5\\
3 & 4 & 1 & 2 & 5\\
1 & 3 & 5 & 2 & 4\\
1 & 4 & 2 & 3 & 5
\end{bmatrix}
\end{align*}

  \item $m=6$:

\begin{align*}
\begin{bmatrix}
1 & 2 & 3 & 6 & 5 & 4\\
2 & 3 & 4 & 1 & 6 & 5\\
3 & 4 & 5 & 2 & 1 & 6\\
1 & 3 & 6 & 5 & 4 & 2\\
3 & 5 & 2 & 1 & 6 & 4\\
5 & 1 & 4 & 3 & 2 & 6\\
1 & 4 & 3 & 6 & 2 & 5\\
3 & 6 & 5 & 1 & 2 & 4\\
1 & 2 & 5 & 3 & 4 & 6\\
1 & 3 & 5 & 2 & 4 & 6
\end{bmatrix}
\end{align*}

  \item $m=7$:
\begin{align*}
\begin{bmatrix}
1 & 2 & 3 & 4 & 5 & 6 & 7\\
2 & 3 & 4 & 7 & 1 & 5 & 6\\
3 & 4 & 7 & 6 & 2 & 1 & 5\\
4 & 7 & 6 & 2 & 5 & 1 & 3\\
1 & 4 & 3 & 7 & 6 & 2 & 5\\
5 & 7 & 4 & 3 & 2 & 1 & 6\\
3 & 7 & 5 & 6 & 1 & 2 & 4\\
2 & 7 & 4 & 1 & 6 & 3 & 5\\
2 & 6 & 3 & 7 & 1 & 4 & 5\\
1 & 7 & 3 & 5 & 2 & 4 & 6\\
3 & 6 & 7 & 1 & 2 & 4 & 5\\
1 & 4 & 7 & 6 & 2 & 3 & 5\\
1 & 2 & 7 & 3 & 4 & 5 & 6\\
1 & 5 & 7 & 2 & 3 & 4 & 6\\
1 & 6 & 7 & 2 & 3 & 4 & 5\\
2 & 3 & 7 & 5 & 1 & 4 & 6\\
2 & 5 & 7 & 4 & 3 & 6 & 1\\
2 & 6 & 7 & 1 & 3 & 4 & 5
\end{bmatrix}
\end{align*}

\end{itemize}

\section{Proof of Theorem~\ref{thm:ub_partitioning_functions_decision_trees}}
\label{app:proof_of_ub_partitioning_functions}

Theorem~\ref{thm:ub_partitioning_functions_decision_trees} relies on a proposition we expose in the following section, and we proceed with the proof thereafter.

\subsection{Formalizing decision trees as partitioning machines}
\label{ssec:formalizing_decision_trees_as_partitioning_machines}

In Section~\ref{sec:paritions_as_a_framework}, we introduce the notion of trees as partitioning machines.
We here formalize this idea by providing a recursive construction of partitions realizable by a tree class $T$.

\begin{figure}[h!]
\centering
\begin{tikzpicture}

\node[draw, circle, minimum width=.5cm](root) {};
\node[above] at (root.north) {root};
\path (root) +(-130:2.35) node[draw, regular polygon, regular polygon sides=3, minimum height=1cm](TL) {\phantom{$T$}};
\node at (TL) {$T_l$};
\path (root) +(-50:2.35) node[draw, regular polygon, regular polygon sides=3, minimum height=1cm](TR) {\phantom{$T$}};
\node at (TR) {$T_r$};
\draw (root) -- node[pos=.5, above, sloped](lambda){\phantom{$\lambda$}} (TL.north);
\node at (lambda) {$\lambda$};
\draw (root) -- node[pos=.5, above, sloped](S-lambda){\phantom{$\lambda$}} (TR.north);
\node at ([xshift=2mm]S-lambda) {$S\backslash\lambda$};

\node[below] at (TL.south) {$\parti{\alpha}$};
\node[below] at (TR.south) {$\parti{\beta}$};

\end{tikzpicture}
\caption{The root node splits the set $S$ into two parts, $\lambda$ and $S\backslash\lambda$, which are forwarded to the left subtree class $T_l$ and the right subtree class $T_r$ respectively. The subtrees produces partitions $\parti{\alpha}$ and $\parti{\beta}$, which can be combined to yield a $c$-partition $\parti{\gamma}$.}
\label{fig:partition_tree}
\end{figure}
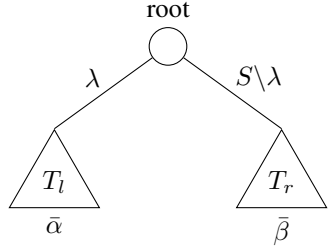

Given a tree class $T$ and a sample $S$, let $\parti{\gamma} \eqdef \cb{\gamma_1,\dots, \gamma_c} \in \P_T^c(S)$ be some $c$-partition realizable by $T$ and let $\parti{\lambda} \eqdef \cb{\lambda, S\backslash\lambda} \in \R(S)$ be a 2-partition realized by the root node which led to $\parti{\gamma}$.
According to our definition~\ref{def:binary_decision_tree} of a binary tree, we have that $\lambda$ is forwarded to the left subtree class $T_l$, which produces an $a$-partition $\parti{\alpha}(\lambda)$ while $S\backslash\lambda$ is sent to the right subtree class $T_r$ which produces a $b$-partition $\parti{\beta}(S\backslash\lambda)$, as pictured in Figure~\ref{fig:partition_tree}.
As explained in Section~\ref{sec:paritions_as_a_framework}, $\parti{\gamma}$ arises from the union of some of the leaves, therefore it also arises from the union of some of the parts in $\parti{\alpha}$ and $\parti{\beta}$.
% By the existence of $\parti{\gamma}$, we have that $\parti{\lambda}, \parti{\alpha}$, and $\parti{\beta}$ exist.
Moreover, this implies that $a+b$ must be greater or equal to $c$.

Note that, generally, there exists multiple partitions $\parti{\alpha}$, $\parti{\beta}$, and $\parti{\lambda}$ that yield the same partition $\parti{\gamma}$.
As a consequence, we can also assume without loss of generality that $a \le c$ and $b\le c$.
Indeed, by construction, any part $\gamma_j \in \parti{\gamma}$ is the result of the union of some subset of parts $\parti{\alpha}^j \subseteq \parti{\alpha}$ and some other subset of parts $\parti{\beta}^j \subseteq \parti{\beta}$.
Note that $\parti{\alpha}^j$ and $\parti{\beta}^j$ can be empty, but not both at the same time.
Using this notation, we have that $\gamma_j = \bigcup_{\alpha \in \parti{\alpha}^j} \alpha \cup \bigcup_{\beta \in \parti{\beta}^j} \beta$ for every $\gamma_j$.
Consider the following partition $\parti{\alpha}'\eqdef \big\{ \alpha_j' : \alpha_j' \eqdef \bigcup_{\alpha \in \parti{\alpha}^j} \alpha, \alpha_j' \neq \varnothing \big\}$ and define $\parti{\beta}'$ similarly.
In this formulation, $\gamma_j$ is equal to $\alpha_j'$, $\beta_j'$, or $\alpha_j' \cup \beta_j'$.
Moreover, the way $\parti{\alpha}'$ and $\parti{\beta}'$ are defined implies that $\parti{\alpha}'\in \P^{a'}_{T_l}(\lambda)$ and $\parti{\beta}'\in \P^{b'}_{T_r}(S\backslash\lambda)$ for $a' \eqdef \abs{\parti{\alpha}'}$ and $b' \eqdef \abs{\parti{\beta}'}$.
Finally, this also implies that $a', b' \leq c$, as wanted.

Having now described how a realizable partition $\parti{\gamma} \in \P_T^c(S)$ of a tree class $T$ is related to the realizable partitions $\parti{\alpha}$ and $\parti{\beta}$ of its left and right subtrees, it is relevant to ask instead what partitions $\parti{\gamma}$ can be made given the partitions $\parti{\alpha}$ and $\parti{\beta}$. 
To do so, we define the following quantity.

\begin{definition}[$c$-partitions-set of pairwise unions of two partitions]\label{def:partitions_set_parwise_unions}
Let $\parti{\alpha}$ be an $a$-partition of some set $A$ and $\parti{\beta}$ be a $b$-partition of some other set $B$, disjoint from $A$.
Define the set $\Q^c(\parti{\alpha}, \parti{\beta})$ of $c$-partitions that can be constructed from pairwise unions of $\parti{\alpha}$ and $\parti{\beta}$ as follows:
\begin{align}
    \Q^c(\parti{\alpha}, \parti{\beta}) \eqdef \left\{ \right. \parti{\gamma} :&\, \parti{\gamma} \text{ is a $c$-partition of $A\cup B$ s.t.} \,\forall\, \gamma \in \parti{\gamma}, \exists\, \alpha \in \parti{\alpha}, \beta \in \parti{\beta} \nonumber\\
    &\text{ s.t.} \,\gamma = \alpha \text{ or } \gamma = \beta \text{ or } \gamma = \alpha \cup \beta \left. \right\}.
\end{align}
\end{definition}

From this definition, it follows that $\Q^c(\parti{\alpha}, \parti{\beta}) = \varnothing$ if $a + b < c$, $a > c$, or $b > c$.
Moreover, if $\mathcal{A}^a(A)$ is some set of $a$-partitions of $A$ and $\mathcal{B}^b(B)$ is some set of $b$-partitions of $B$, we denote by
\begin{equation*}
    \Q^c(\mathcal{A}^a(A), \mathcal{B}^b(B)) \eqdef \bigcup_{\substack{\parti{\alpha}\in\A^a(A),\\\parti{\beta}\in\mathcal{B}^b(B)}} \Q^c(\parti{\alpha}, \parti{\beta})
\end{equation*}
the union set of the $\Q^c$.

We are now equipped to write a recursive relation of the set of partitions a tree $T$ can realize knowing the set of partitions its subtrees can realize.

\begin{proposition}[$c$-partitions-set decomposition of decision trees]
\label{prop:c-partitions-set_decomposition_decision_trees}
Let $\P_T^c(S)$ be the set of $c$-partitions that a binary decision tree class $T$ can realize on a sample $S$ of $m > L_T$ examples, and let $T_l$ and $T_r$ be the hypothesis classes of its left and right subtrees.
Then, the following decomposition holds.
\begin{equation}\label{eq:c-partitions-set_decomposition}
    \P^c_T(S) = \!\!\!\bigcup_{\cb{\lambda, S\backslash\lambda} \in \R(S)} \;\bigcup_{1 \leq a, b \leq c}  \Q^c\pr{ \P^a_{T_l}(\lambda), \P^b_{T_r}(S\backslash\lambda) } \cup \Q^c\pr{ \P^a_{T_l}(S\backslash\lambda), \P^b_{T_r}(\lambda) },
\end{equation}
where $\R(S)$ denotes the set of 2-partitions the root node can realize on $S$.
\end{proposition}

\begin{proof}
Let $\parti{\gamma} \in \P^c_T(S)$.
Then, our explanations in the paragraphs above Definition~\ref{def:partitions_set_parwise_unions} imply that if $\lambda$ is forwarded to $T_l$, then $\parti{\gamma} \in \Q^c ( \P^a_{T_l}(\lambda), \P^b_{T_r}(S\backslash\lambda) )$.
Alternatively, if $\lambda$ is forwarded to $T_r$, then a similar reasoning shows that $\Q^c (\P^a_{T_l}(S\backslash\lambda), \P^b_{T_r}(\lambda) )$.
On the other hand, we also have discussed that $\Q^c(\parti{\alpha}, \parti{\beta})$ is the quantity that contains all $c$-partitions realizable from $\parti{\alpha}$ and $\parti{\beta}$.
Therefore, taking the union over all the partitions realizable by $T_l$ and $T_r$ indeed gives $\P^c_T(S)$.
\end{proof}

\subsection{Proof of the theorem}

We are now ready to prove Theorem~\ref{thm:ub_partitioning_functions_decision_trees}.

\begin{proof}
We consider first the case where the number of examples $m$ is less than or equal to the number of leaves $L_T$ of the tree $T$.
We want to show that there exists a sample $S$ such that $T$ can realize every $c$-partitions of $S$.

Let $S$ be a sample such that one feature takes distinct values for each of the $m$ examples.
Then, one can choose for the root of $T$ the appropriate threshold on that feature such that $m_l \le L_{T_l}$ examples will be redirected to the left and $m_r \le L_{T_r}$ examples will be redirected to the right (where we have $L_{T_l} + L_{T_r} = L_T$ and $m_l + m_r = m$).
Then each of the subtrees can do the required split on the same feature, with the required constraints on the number of examples that need to be redirected on each children, until that we have eventually at most one example per leaf.
In that case, by choosing any labeling in $[c]$ for the leaves, the tree class $T$ can perform any $c$-partition of the $m$ examples out of the $\smallstirling{m}{c}$ possible ones.
Consequently, we have $\pi^c_T(m) = \smallstirling{m}{c}$ for any tree class with $L_T \ge m$.

We now consider the case when $m > L_T$.
We want to show the following inequality:
\begin{equation}\label{eq:app_thm_restatement}
    \pi^c_T(m) \leq \pr{\frac{1}{2}}^{\delta_{lr}} \sum_{k=L_{T_l}}^{m-L_{T_r}} \!\!\min \cb{ 2\ell, \tbinom{m}{k} } \hspace{-7pt} {\sum_{\substack{1 \leq a, b \leq c \\ a + b \geq c}}} \!\!\tbinom{a}{c-b} \tbinom{b}{c-a} (a + b - c)!\; \pi^a_{T_l}(k) \pi^b_{T_r}(m-k)\, ,
\end{equation}
where $\delta_{lr} = 1$ if $T_l = T_r$, and $0$ otherwise.

In the following, we assume every examples of $S$ have distinct feature values, \ie it is always possible to distinguish two examples using any feature.
Indeed, assuming otherwise can only reduce the number of partitions that can be made on a sample, and therefore we have, for such a sample $S$ of $m$ examples, that $\abs{P_T^c(S)} \le \pi^c_T(m)$.

We start from Proposition~\ref{prop:c-partitions-set_decomposition_decision_trees}, which states
\begin{equation}
    \P^c_T(S) = \bigcup_{\cb{\lambda, S\backslash\lambda} \in \R(S)} \;\bigcup_{1 \leq a, b \leq c}  \Q^c\pr{ \P^a_{T_l}(\lambda), \P^b_{T_r}(S\backslash\lambda) } \cup \Q^c\pr{ \P^a_{T_l}(S\backslash\lambda), \P^b_{T_r}(\lambda) }.
\end{equation}
Because $\Q^c$ is symmetric in its arguments and because the union over $a$ and $b$ is invariant under the exchange of $a$ and $b$, we have that
\begin{equation}
    \bigcup_{a,b} \Q^c \pr{\P^a_{T_l}(S\backslash\lambda), \P^b_{T_r}(\lambda)} = \bigcup_{a,b} \Q^c \pr{ \P^a_{T_r}(\lambda), \P^b_{T_l}(S\backslash\lambda) },
\end{equation}
which is equivalent to say that one can exchange the subtrees instead of sending $\lambda$ to the left and to the right.
Therefore, we have
\begin{equation}\label{eq:PcT_as_union_of_Alr_Arl}
    \P^c_T(S) = \A_{lr} \cup \A_{rl} \quad \text{with } \A_{lr} \eqdef \hspace{-13pt} \bigcup_{\cb{\lambda, S\backslash\lambda} \in \R(S)} \;\bigcup_{1 \leq a, b \leq c}  \Q^c\pr{ \P^a_{T_l}(\lambda), \P^b_{T_r}(S\backslash\lambda) },
\end{equation}
where we mute the other dependencies of $\A$ to alleviate the notation.

By the union bound, we have $\abs{\P^c_T(S)} \leq \abs{\A_{lr}} + \abs{\A_{rl}}$. Let us upper bound $\abs{\A_{lr}}$.
Observe that a single node is very similar to a decision stump.
Indeed, the root partitions decomposition of Equation~\eqref{eq:root_decomposition} also applies here, so that we have
\begin{equation}
    \A_{lr} = \bigcup_{k=1}^{\floor{\frac{m}{2}}} \bigcup_{\cb{\lambda, S\backslash\lambda} \in \R_k(S)} \;\bigcup_{1 \leq a, b \leq c}  \Q^c\pr{ \P^a_{T_l}(\lambda), \P^b_{T_r}(S\backslash\lambda) }.
\end{equation}

Then, we show that the union over $k$ can be changed to go from $L_{T_l}$ to $\min\cb{\floor{\frac{m}{2}}\!, m - L_{T_r}}$ without changing $\A_{lr}$.
To do so, we need to show that for any partition $\parti{\gamma} \in \A_{lr}$, there exists at least one 2-partition $\plambda = \{\lambda,S\backslash\lambda\}$ realized by the root node with $L_{T_l} \le \abs{\lambda} \le \min\cb{\floor{\frac{m}{2}}\!, m - L_{T_r}}$ that leads to $\parti{\gamma}$.
Indeed, assume $\abs{\lambda} < L_{T_l}$.
Because of our assumption below Equation~\ref{eq:app_thm_restatement}, one can always modify the threshold of the root node to send $L_{T_l}$ examples in the subtree $T_l$ and modify the subtree so that every example ends up alone in a leaf (as we have shown in the first part of the present proof).
These examples can then be united into the part they belonged in $\parti{\gamma}$ to give the same partition as before.
An analogous argument also holds for $\abs{S\backslash\lambda} \geq L_{T_r}$, which implies $\abs{\lambda} \le m - L_{T_r}$ (since $m > L_T$ by assumption).

Letting $M_r \eqdef \min\cb{\floor{\frac{m}{2}}, m - L_{T_r}}$ and taking the union bound over $k$ and over $\R_k(S)$, one ends up with
\begin{equation}
    \abs{\A_{lr}} \leq \sum_{k=L_{T_l}}^{M_r} \abs{\R_k(S)} \max_{\cb{\lambda, S\backslash\lambda} \in \R_k(S)} \abs{\bigcup_{1 \leq a, b \leq c}  \Q^c\pr{ \P^a_{T_l}(\lambda), \P^b_{T_r}(S\backslash\lambda) }}. \label{eq:cardinality_Alr_from_cardinality_of_Qc}
\end{equation}

Let us evaluate the cardinality of $\Q^c\pr{ \P^a_{T_l}(\lambda), \P^b_{T_r}(S\backslash\lambda) }$ when $1 \leq a, b \leq c$ and $a + b \geq c$.
Using the union bound over disjoint events, we have
\begin{align}
    \abs{\Q^c\pr{ \P^a_{T_l}(\lambda), \P^b_{T_r}(S\backslash\lambda) }} &= \sum_{\parti{\alpha} \in \P^a_{T_l}(\lambda)} \sum_{\parti{\beta} \in \P^b_{T_r}(S\backslash\lambda)} \big|\Q^c\pr{ \parti{\alpha}, \parti{\beta} }\big|.\label{eq:cardinality_Qc_union_bound}
\end{align}
Pick any $\parti{\alpha} \in \P^a_{T_l}(\lambda)$ and $\parti{\beta} \in \P^b_{T_r}(S\backslash\lambda)$.
According to Definition~\ref{def:partitions_set_parwise_unions}, we must take the unions of some parts of $\parti{\alpha}$ and $\parti{\beta}$ to end up with a $c$-partition, with the constraint that the joined parts belongs to different partitions.
We start with a total of $a+b$ parts and we must take the union of some pairs to end up with only $c$ parts.
Taking the union of such a pair effectively reduces the total number of parts by one, therefore we must make $a+b-c$ unions.
To make these unions, choose $a+b-c$ parts from $\parti{\alpha}$ and choose $a+b-c$ parts from $\parti{\beta}$ and join them.
Since there is $(a+b-c)!$ ways to join those parts, we have that $\abs{\Q^c(\parti{\alpha}, \parti{\beta})} = \binom{a}{c-b} \binom{b}{c-a} (a + b - c)!$.
Since the cardinality of $\Q^c(\parti{\alpha}, \parti{\beta})$ depends only on $a$ and $b$ and not the partitions themselves, Equation~\eqref{eq:cardinality_Qc_union_bound} becomes
\begin{align}\label{eq:cardinality_Qc_Pa_Pb}
    \abs{\Q^c\pr{ \P^a_{T_l}(\lambda), \P^b_{T_r}(S\backslash\lambda) }} = \tbinom{a}{c-b} \tbinom{b}{c-a} (a+b-c)! \big|\P^a_{T_l}(\lambda)\big| \big|\P^b_{T_r}(S\backslash\lambda)\big|.
\end{align}

Going back to Equation~\eqref{eq:cardinality_Alr_from_cardinality_of_Qc}, one has
\begin{align}
    \abs{\A_{lr}} &\leq \sum_{k=L_{T_l}}^{M_r} \abs{\R_k(S)} \hspace{-4pt}\sum_{\substack{1 \leq a, b, \leq c \\ a+b \geq c}}\hspace{-4pt} \tbinom{a}{c-b} \tbinom{b}{c-a} (a + b - c)! \max_{\cb{\lambda, S\backslash\lambda} \in \R_k(S)} \big|\P^a_{T_l}(\lambda)\big| \big|\P^b_{T_r}(S\backslash\lambda)\big|.
\end{align}
Then, using Definition~\ref{def:partitioning_function} for $\pi^c_T(m)$ yields
\begin{align}
    \abs{\A_{lr}}\leq \sum_{k=L_{T_l}}^{M_r} \abs{\R_k(S)} \hspace{-4pt}\sum_{\substack{1 \leq a, b, \leq c \\ a+b \geq c}}\hspace{-4pt} \tbinom{a}{c-b} \tbinom{b}{c-a} (a + b - c)!\; \pi^a_{T_l}(k) \pi^b_{T_r}(m-k).\label{eq:Alr_bound_1}
\end{align}

This expression also applies to $\A_{rl}$ by exchanging indices $l$ and $r$.
Apply this exchange to Equation~\eqref{eq:Alr_bound_1}. Then let $k \to m-k$, and rename $a$ to $b$ and $b$ to $a$, so that we have
\begin{align}
    \abs{\A_{rl}} \leq \sum_{k=M_l}^{m-L_{T_r}} \abs{\R_k(S)} \hspace{-4pt}\sum_{\substack{1 \leq a, b, \leq c \\ a+b \geq c}}\hspace{-4pt} \tbinom{a}{c-b} \tbinom{b}{c-a} (a + b - c)!\; \pi^a_{T_l}(k) \pi^b_{T_r}(m-k),\label{eq:Arl_bound}
\end{align}
where $M_l \eqdef \max\cb{\ceil{\frac{m}{2}}\!, L_{T_l} }$.
Notice that the coefficients inside the sum over $k$ are the same in Equations~\eqref{eq:Alr_bound_1} and \eqref{eq:Arl_bound}.
For convenience, let
\begin{equation}
    C_k \eqdef \sum_{\substack{1 \leq a, b, \leq c \\ a+b \geq c}}\hspace{-4pt} \tbinom{a}{c-b} \tbinom{b}{c-a} (a + b - c)!\; \pi^a_{T_l}(k) \pi^b_{T_r}(m-k),
\end{equation}
so that $\abs{\A_{lr}}$ and $\abs{\A_{rl}}$ can written in the form $\sum_k \abs{\R_k(S)} C_k$, with the only difference being the values that $k$ takes.
We can now show that the sum over $k$ in Equations~\eqref{eq:Alr_bound_1} and \eqref{eq:Arl_bound} can be put together to yield the theorem.

There are 4 cases to consider according to the values of $M_r = \min\cb{\floor{\frac{m}{2}}, m - \Lr}$ and $M_l = \max\cb{\ceil{\frac{m}{2}}, L_{T_l}}$.
First, let $M_r = \floor{\frac{m}{2}}$ and $M_l = \ceil{\frac{m}{2}}$.
The sum over $k$ then goes from $L_{T_l}$ to $\floor{\frac{m}{2}}$ for $\abs{\A_{lr}}$ and from $\ceil{\frac{m}{2}}$ to $m-L_{T_r}$ for $\abs{\A_{rl}}$.
Then, if $m$ is odd, both sums can be joined directly to go from $L_{T_l}$ to $m-L_{T_{r}}$.
If $m$ is even, one has an extra term for $k = \frac{m}{2}$. Thus
\begin{equation}
    \abs{\A_{lr}} + \abs{\A_{rl}} \leq \left\{ \begin{array}{ll}\displaystyle
         \sum_{k=L_{T_l}}^{m-L_{T_r}} \abs{\R_k(S)} C_k & \text{if $m$ is odd}\\
         \displaystyle
        \abs{\R_{\frac{m}{2}}(S)} C_{\frac{m}{2}} + \sum_{k=L_{T_l}}^{m-L_{T_r}} \abs{\R_k(S)} C_k & \text{if $m$ is even.}
    \end{array}\right.
\end{equation}
Using the upper bound on $\abs{\R_k(S)}$ in Equation~\eqref{eq:bound_on_RkS}, the above expression simplifies to
\begin{equation}
    \abs{\A_{lr}} + \abs{\A_{rl}} \leq \sum_{k=L_{T_l}}^{m-L_{T_r}} \min\cb{2\ell, \tbinom{m}{k}} C_k,
\end{equation}
valid for both cases.

Second, let $M_r = \min\cb{\floor{\frac{m}{2}}, m - \Lr} = \floor{\frac{m}{2}}$ and $M_l = \max\cb{\ceil{\frac{m}{2}}, L_{T_l}} = L_{T_l}$.
This implies that $L_{T_l} \geq \ceil{\frac{m}{2}}$.
The sum over $k$ then goes from $L_{T_l}$ to $\floor{\frac{m}{2}}$ for $\abs{\A_{lr}}$, which consists in exactly one term if $L_{T_l}=\frac{m}{2}$ and none otherwise.
For $\abs{\A_{rl}}$, the sum over $k$ goes from $\Ll$ to $m-\Lr$.
Therefore, we have
\begin{equation}
    \abs{\A_{lr}} + \abs{\A_{rl}} \leq \left\{ \begin{array}{ll}
        \displaystyle
        \abs{\R_{\frac{m}{2}}(S)} C_{\frac{m}{2}} + \sum_{k=L_{T_l}}^{m-L_{T_r}} \abs{\R_k(S)} C_k & \text{if $\Ll = \frac{m}{2}$.}\\
        \displaystyle
        \sum_{k=L_{T_l}}^{m-L_{T_r}} \abs{\R_k(S)} C_k & \text{otherwise}
    \end{array}\right.
\end{equation}
Again, using the upper bound on $\abs{\R_k(S)}$ in Equation~\eqref{eq:bound_on_RkS}, the above expression simplifies to
\begin{equation}
    \abs{\A_{lr}} + \abs{\A_{rl}} \leq \sum_{k=L_{T_l}}^{m-L_{T_r}} \min\cb{2\ell, \tbinom{m}{k}} C_k,
\end{equation}
valid for both cases.

Third, let $M_r = \min\cb{\floor{\frac{m}{2}}, m - \Lr} = m-\Lr$ and $M_l = \max\cb{\ceil{\frac{m}{2}}, L_{T_l}} = \ceil{\frac{m}{2}}$.
This case is very similar to the second case, where $\abs{\A_{rl}}$ consists in one or zero term instead of $\abs{\A_{lr}}$.
Thus, the same conclusion applies.

Fourth, let $M_r = \min\cb{\floor{\frac{m}{2}}, m - \Lr} = m-\Lr$ and $M_l = \max\cb{\ceil{\frac{m}{2}}, L_{T_l}} = \Ll$.
This case violates our starting assumption that $m$ is greater than $L_T$.
Hence, we can simply ignore this case.

Collecting our results, one concludes that for all $m > \Ll + \Lr$, we have
\begin{equation}\label{eq:almost_bound_on_picT}
    \abs{\P^c_T(S)} \leq \abs{\A_{lr}} + \abs{\A_{rl}} \leq \sum_{k=L_{T_l}}^{m-L_{T_r}} \min\cb{2\ell, \tbinom{m}{k}} C_k.
\end{equation}
Observe that the right-hand-side of this inequality is independent of $S$.
Therefore, by taking the maximum value over all sample $S$ of size $m$, we have a bound for $\pi^c_T(m)$.

One can improve this result when the left and the right subtrees are the same.
Indeed, in this case $\A_{lr} = \A_{rl}$ so that $\P^c_T(S)$ is simply equal to $\A_{lr}$ according to Equation~\eqref{eq:PcT_as_union_of_Alr_Arl}.
Moreover, the condition that $m  > \Ll + \Lr$ implies $\Lr < \frac{m}{2}$, so that $M_r$ is always equal to $\floor{\frac{m}{2}}$.
Equation~\eqref{eq:Alr_bound_1} then becomes
\begin{align}
    \abs{\A_{lr}} \leq \sum_{k=\Ll}^{\floor{\frac{m}{2}}} \abs{\R_k(S)} \hspace{-4pt}\sum_{\substack{1 \leq a, b, \leq c \\ a+b \geq c}}\hspace{-4pt} \tbinom{a}{c-b} \tbinom{b}{c-a} (a + b - c)!\; \pi^a_{T_l}(k) \pi^b_{T_r}(m-k).
\end{align}
Using the fact that $\R_k(S) = \R_{m-k}(S)$, that $T_l = T_r$, and that the summation over $a$ and $b$ is symmetric, along with the bound of Equation~\eqref{eq:bound_on_RkS} on $\abs{\R_k(S)}$, one can show that
\begin{align}
    \abs{\P^c_T(S)} \leq \abs{\A_{lr}} \leq \frac{1}{2}\sum_{k=\Ll}^{m-\Lr} \min\cb{2\ell, \tbinom{m}{k}} \hspace{-4pt}\sum_{\substack{1 \leq a, b, \leq c \\ a+b \geq c}}\hspace{-4pt} \tbinom{a}{c-b} \tbinom{b}{c-a} (a + b - c)!\; \pi^a_{T_l}(k) \pi^b_{T_r}(m-k),
\end{align}
which is different from Equation~\eqref{eq:almost_bound_on_picT} by a factor of $1/2$ only.

We finally obtain the statement of the theorem if we use the indicator function $\mathds{1}[\cdot]$ to handle into a single expression the cases when $T_l$ and $T_r$ are the same or not.
\end{proof}

\clearpage
\section{Proof of Corollary~\ref{coro:asymptotic_VC_tree}}
\label{app:proof_coro_asymptotic_vcdim}
We here give the proof of Corollary~\ref{coro:asymptotic_VC_tree}, which states that the asymptotic behavior of the VC dimension of a class $T$ of a binary decision tree with $N$ internal nodes on examples of $\ell$ real-valued features is given by $\vcdim T \in O\pr{ N \log(N\ell) }$.

\begin{proof}
Letting $c=2$ in Theorem~\ref{thm:ub_partitioning_functions_decision_trees}, using the fact that $2^{-\delta_{lr}} \le 1$, $\min\cb{2\ell,\binom{m}{k}} \le 2\ell$ and $\pi^c_{T}(k) \le \pi^c_T(m)$ for $k \le m$, we have
\begin{equation*}
    \pi^2_T(m) \le 2\ell (m-L_T) \pr{1 + 2\pi^2_{T_l}(m) + 2\pi^2_{T_r}(m) +2\pi^2_{T_l}(m)\pi^2_{T_r}(m) }.
\end{equation*}
We show by induction that $\pi^2_T(m) \in O((m\ell)^N)$.
Assume $\pi^2_T(m) \le (Cm\ell)^N$ for some constant $C\ge 1$, and let $N_l$ and $N_r$ be the number of nodes in the left and right subtrees respectively, so that $N_l+N_r+1=N$.
The previous equation becomes (with $m-L_T<m$)
\begin{align*}
    \pi^c_T(m) &\le 2m\ell\pr{1 + 2(Cm\ell)^{N_l} + 2(Cm\ell)^{N_r} + 2(Cm\ell)^{N_l} (Cm\ell)^{N_r}}\\
    &\le 14m\ell (Cm\ell)^{N_l+N_r},
\end{align*}
which proves our claim for $C \ge 14$.
Then, Equation~\eqref{eq:ndim_def_partition_func} implies
\begin{equation*}
    \vcdim T \le \max \cb{ m : (Cm\ell)^N \ge 2^{m-1}-1}.
\end{equation*}
One can solve for the inequality $(Cm\ell)^N \ge 2^m$ instead, since this implies $(Cm\ell)^N \ge 2^{m-1}-1$ is true too.
The Lambert $W$ function \citep{corless96lambertw} can give us an exact solution, which is $m \le - \frac{N}{\ln 2} W_{-1} \pr{ -\frac{\ln 2}{C N \ell} }$.
Since $-W_{-1}(-z^{-1}) \in O\pr{\log z}$, we have that $\vcdim T \in O\pr{N\log(N\ell)}$.

\end{proof}

\clearpage
\section{Algorithms to upper bound the VC dimension of decision tree classes}
\label{app:algorithms_bound_vcdim_trees}

In this Appendix, we present the algorithms for obtaining an upper bound on the VC dimension of a tree class $T$.
Algorithm~\ref{algo:partition_func_upper_bound} uses  Theorem~\ref{thm:ub_partitioning_functions_decision_trees} to upper bound the $c$-partitioning function of a tree class.
Algorithm~\ref{algo:vcdim_upper_bound} uses Algorithm~\ref{algo:partition_func_upper_bound} and Equation~\eqref{eq:ndim_def_partition_func} to compute an upper bound on the VC dimension of a tree class.
% Algorithm~\ref{algo:vcdim_lower_bound} uses Theorem~\ref{thm:lb_vcdim_decision_trees} and Corollary~\ref{thm:vcdim_stump} (when it reaches a node which is a decision stump) to compute a lower bound on the VC dimension of a tree class.

\begin{algorithm}
\caption{PartitionFuncUpperBound$(T, c, m, \ell)$}\label{algo:partition_func_upper_bound}
\DontPrintSemicolon
\SetAlgoVlined

\KwIn{A tree class $T$, the number $c$ of parts in the partitions, the number $m$ of elements, the number $\ell$ of features.}
Let $L_T$ be the number of leaves of $T$.\;

\uIf{$c > m$ or $c > L_T$}{
	Let $N \leftarrow 0$.\;
	}
\uElseIf{$c = m$ or $c = 1$ or $m = 1$}{
	Let $N \leftarrow 1$.\;
	}
\uElseIf{$m \leq L_T$}{
    Let $N \leftarrow \smallstirling{m}{c}$.}
\Else{
	Let $T_l$ and $T_r$ be the left and right subtree classes of $T$.\;
	
	Let $N \leftarrow 0$.\;
	
	\For{$k = \Ll$ to $m-\Lr$}{
	   % Let $M \leftarrow 0$.\;
	    
	   % \For{$a = 1$ to $c$}{
	   %     \For{$b=c-a+1$ to $c$}{
	   %         \raggedright $M \leftarrow M + \binom{a}{c-b}\binom{b}{c-a}(a+b-c)! \cdot \text{PartitionFuncUpperBound}(T_l, a, k, \ell)$\;
	   %         \hspace*{5.4cm}$\cdot\; \text{PartitionFuncUpperBound}(T_r, b, m-k, \ell)$.\;
	   %     }
        % }
        \raggedright Let $\displaystyle N \leftarrow N + \min\cb{2\ell, \tbinom{m}{k}} \sum_{a=1}^c \sum_{b=\max\cb{1,c-a}}^c \tbinom{a}{c-b}\tbinom{b}{c-a}(a+b-c)!$\;
            \hspace*{5.9cm}$\times\;\text{PartitionFuncUpperBound}(T_l, a, k, \ell)$\;
            \hspace*{5.9cm}$\times\; \text{PartitionFuncUpperBound}(T_r, b, m-k, \ell)$.\;
	            
        % Let $N \leftarrow N + \min\cb{2\ell, \binom{m}{k}} \cdot M$.\;
        
        \If{$T_L = T_R$}{
            Let $N \leftarrow \frac{N}{2}$.\;
        }
	}
}

\KwOut{$\min\pr{N, \smallstirling{m}{c}}$.}
\end{algorithm}

\begin{algorithm}[ht]
\caption{VCdimUpperBound$(T, \ell)$}\label{algo:vcdim_upper_bound}
\DontPrintSemicolon
\SetAlgoVlined

\KwIn{A tree class $T$, the number $\ell$ of features.}

\If{$T$ is a leaf}{\KwOut{1}}

Let $m \leftarrow L_T+1$.\;

\While{$\text{\upshape PartitionFuncUpperBound}(T, 2, m, \ell) \geq 2^{m-1}-1$}{
    Let $m \leftarrow m+1.$\;
	}
\KwOut{$m-1$}
\end{algorithm}

Algorithm~\ref{algo:vcdim_upper_bound} can become quite inefficient because one has to compute the values of PartitionFuncUpperBound for increasing values of $m$, which may already have been computed for smaller values of $m$.
It is thus suggested to store the values of PartitionFuncUpperBound computed for each $T$ and each $m$ to be more efficient.

% \begin{algorithm}[ht]
% \caption{VCdimLowerBound$(T,\ell)$}\label{algo:vcdim_lower_bound}
% \DontPrintSemicolon
% \SetAlgoVlined

% \KwIn{A tree class $T$, the number $\ell$ of features.}

% \If{$T$ is a leaf}{\KwOut{1}}

% \If(\tcp*[h]{VCdimUpperBound is exact for a stump}){$T$ is a stump}{\KwOut{$\text{VCdimUpperBound}(T, \ell)$}}

% Let $T_l$ and $T_r$ be the left and right subtree classes of $T$.\;

% \KwOut{$\text{VCdimLowerBound}(T_l, \ell) + \text{VCdimLowerBound}(T_r, \ell)$}
% \end{algorithm}

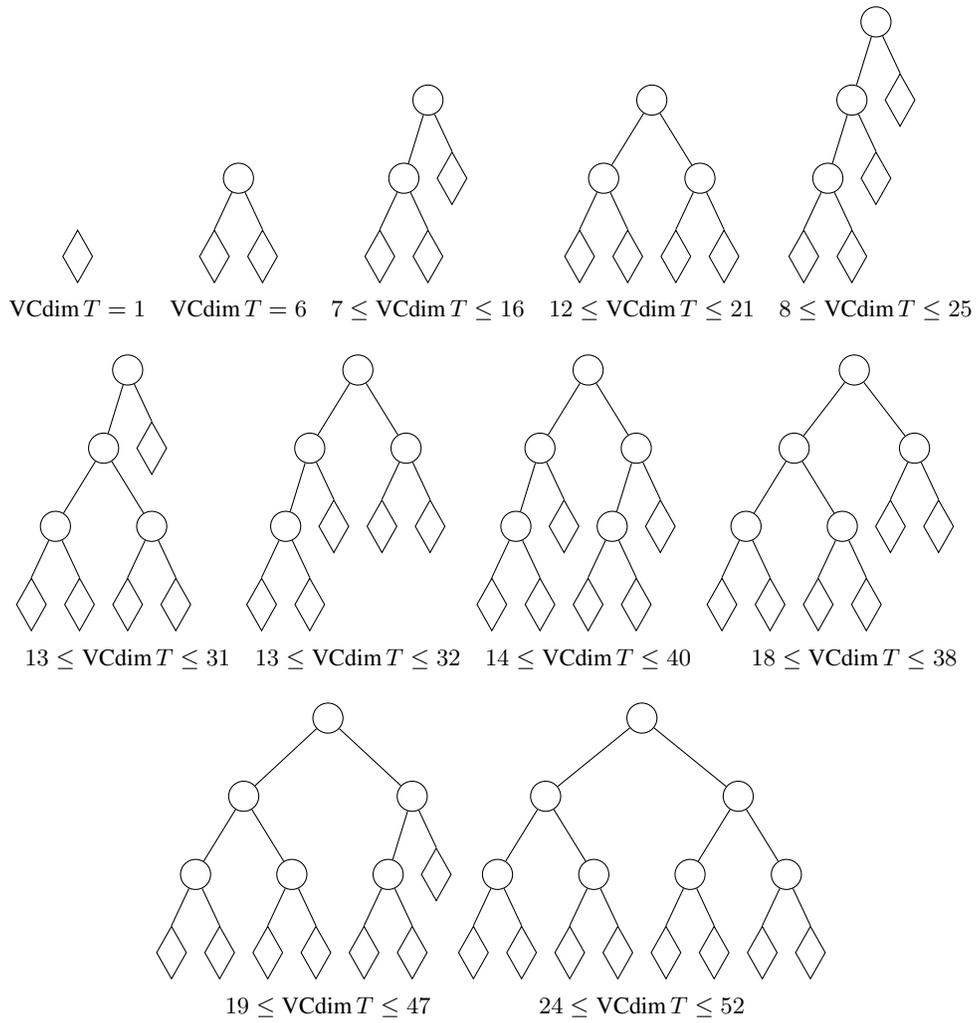
\begin{figure}[h!]
\centering
\begin{tikzpicture}[leaf/.style={draw, diamond, minimum width=0.4cm, aspect=.5}, internal/.style={draw, circle, minimum width=0.4cm}, scale=.8]
\node[leaf](node0) at (0, 1.3) {};
\node(node1) at (0, 0.4) {\footnotesize $\textrm{VCdim}\,T = 1\vphantom{\leq}$};
\end{tikzpicture}
\begin{tikzpicture}[leaf/.style={draw, diamond, minimum width=0.4cm, aspect=.5}, internal/.style={draw, circle, minimum width=0.4cm}, scale=.8]
\node[internal](node0) at (0, 2.6) {};
\node[leaf](node1) at (-0.4, 1.3) {};
\draw (node1.north) -- (node0);
\node[leaf](node2) at (0.4, 1.3) {};
\draw (node2.north) -- (node0);
\node(node2) at (0, 0.4) {\footnotesize $\textrm{VCdim}\,T = 6\vphantom{\leq}$};
\end{tikzpicture}
\begin{tikzpicture}[leaf/.style={draw, diamond, minimum width=0.4cm, aspect=.5}, internal/.style={draw, circle, minimum width=0.4cm}, scale=.8]
\node[internal](node0) at (0, 3.9000000000000004) {};
\node[internal](node1) at (-0.4, 2.6) {};
\draw (node1) -- (node0);
\node[leaf](node2) at (0.4, 2.6) {};
\draw (node2.north) -- (node0);
\node[leaf](node3) at (-0.8, 1.3) {};
\draw (node3.north) -- (node1);
\node[leaf](node4) at (0.0, 1.3) {};
\draw (node4.north) -- (node1);
\node(node3) at (0, 0.4) {\footnotesize $7 \leq \textrm{VCdim}\,T \leq 16$};
\end{tikzpicture}
\begin{tikzpicture}[leaf/.style={draw, diamond, minimum width=0.4cm, aspect=.5}, internal/.style={draw, circle, minimum width=0.4cm}, scale=.8]
\node[internal](node0) at (0, 3.9000000000000004) {};
\node[internal](node1) at (-0.8, 2.6) {};
\draw (node1) -- (node0);
\node[internal](node2) at (0.8, 2.6) {};
\draw (node2) -- (node0);
\node[leaf](node3) at (-1.2000000000000002, 1.3) {};
\draw (node3.north) -- (node1);
\node[leaf](node4) at (-0.4, 1.3) {};
\draw (node4.north) -- (node1);
\node[leaf](node5) at (0.4, 1.3) {};
\draw (node5.north) -- (node2);
\node[leaf](node6) at (1.2000000000000002, 1.3) {};
\draw (node6.north) -- (node2);
\node(node4) at (0, 0.4) {\footnotesize $12 \leq \textrm{VCdim}\,T \leq 21$};
\end{tikzpicture}
\begin{tikzpicture}[leaf/.style={draw, diamond, minimum width=0.4cm, aspect=.5}, internal/.style={draw, circle, minimum width=0.4cm}, scale=.8]
\node[internal](node0) at (0, 5.2) {};
\node[internal](node1) at (-0.4, 3.9000000000000004) {};
\draw (node1) -- (node0);
\node[leaf](node2) at (0.4, 3.9000000000000004) {};
\draw (node2.north) -- (node0);
\node[internal](node3) at (-0.8, 2.6) {};
\draw (node3) -- (node1);
\node[leaf](node4) at (0.0, 2.6) {};
\draw (node4.north) -- (node1);
\node[leaf](node5) at (-1.2000000000000002, 1.3) {};
\draw (node5.north) -- (node3);
\node[leaf](node6) at (-0.4, 1.3) {};
\draw (node6.north) -- (node3);
\node(node5) at (0, 0.4) {\footnotesize $8 \leq \textrm{VCdim}\,T \leq 25$};
\end{tikzpicture}

\vspace{3mm}

\begin{tikzpicture}[leaf/.style={draw, diamond, minimum width=0.4cm, aspect=.5}, internal/.style={draw, circle, minimum width=0.4cm}, scale=.8]
\node[internal](node0) at (0, 5.2) {};
\node[internal](node1) at (-0.4, 3.9000000000000004) {};
\draw (node1) -- (node0);
\node[leaf](node2) at (0.4, 3.9000000000000004) {};
\draw (node2.north) -- (node0);
\node[internal](node3) at (-1.2000000000000002, 2.6) {};
\draw (node3) -- (node1);
\node[internal](node4) at (0.4, 2.6) {};
\draw (node4) -- (node1);
\node[leaf](node5) at (-1.6, 1.3) {};
\draw (node5.north) -- (node3);
\node[leaf](node6) at (-0.8, 1.3) {};
\draw (node6.north) -- (node3);
\node[leaf](node7) at (0.0, 1.3) {};
\draw (node7.north) -- (node4);
\node[leaf](node8) at (0.8, 1.3) {};
\draw (node8.north) -- (node4);
\node(node6) at (0, 0.4) {\footnotesize $13 \leq \textrm{VCdim}\,T \leq 31$};
\end{tikzpicture}
\begin{tikzpicture}[leaf/.style={draw, diamond, minimum width=0.4cm, aspect=.5}, internal/.style={draw, circle, minimum width=0.4cm}, scale=.8]
\node[internal](node0) at (0, 5.2) {};
\node[internal](node1) at (-0.8, 3.9000000000000004) {};
\draw (node1) -- (node0);
\node[internal](node2) at (0.8, 3.9000000000000004) {};
\draw (node2) -- (node0);
\node[internal](node3) at (-1.2000000000000002, 2.6) {};
\draw (node3) -- (node1);
\node[leaf](node4) at (-0.4, 2.6) {};
\draw (node4.north) -- (node1);
\node[leaf](node5) at (0.4, 2.6) {};
\draw (node5.north) -- (node2);
\node[leaf](node6) at (1.2000000000000002, 2.6) {};
\draw (node6.north) -- (node2);
\node[leaf](node7) at (-1.6, 1.3) {};
\draw (node7.north) -- (node3);
\node[leaf](node8) at (-0.8, 1.3) {};
\draw (node8.north) -- (node3);
\node(node7) at (0, 0.4) {\footnotesize $13 \leq \textrm{VCdim}\,T \leq 32$};
\end{tikzpicture}
\begin{tikzpicture}[leaf/.style={draw, diamond, minimum width=0.4cm, aspect=.5}, internal/.style={draw, circle, minimum width=0.4cm}, scale=.8]
\node[internal](node0) at (0, 5.2) {};
\node[internal](node1) at (-0.8, 3.9000000000000004) {};
\draw (node1) -- (node0);
\node[internal](node2) at (0.8, 3.9000000000000004) {};
\draw (node2) -- (node0);
\node[internal](node3) at (-1.2000000000000002, 2.6) {};
\draw (node3) -- (node1);
\node[leaf](node4) at (-0.4, 2.6) {};
\draw (node4.north) -- (node1);
\node[internal](node5) at (0.4, 2.6) {};
\draw (node5) -- (node2);
\node[leaf](node6) at (1.2000000000000002, 2.6) {};
\draw (node6.north) -- (node2);
\node[leaf](node7) at (-1.6, 1.3) {};
\draw (node7.north) -- (node3);
\node[leaf](node8) at (-0.8, 1.3) {};
\draw (node8.north) -- (node3);
\node[leaf](node9) at (0.0, 1.3) {};
\draw (node9.north) -- (node5);
\node[leaf](node10) at (0.8, 1.3) {};
\draw (node10.north) -- (node5);
\node(node8) at (0, 0.4) {\footnotesize $14 \leq \textrm{VCdim}\,T \leq 40$};
\end{tikzpicture}
\begin{tikzpicture}[leaf/.style={draw, diamond, minimum width=0.4cm, aspect=.5}, internal/.style={draw, circle, minimum width=0.4cm}, scale=.8]
\node[internal](node0) at (0, 5.2) {};
\node[internal](node1) at (-1.0, 3.9000000000000004) {};
\draw (node1) -- (node0);
\node[internal](node2) at (1.0, 3.9000000000000004) {};
\draw (node2) -- (node0);
\node[internal](node3) at (-1.8000000000000003, 2.6) {};
\draw (node3) -- (node1);
\node[internal](node4) at (-0.20000000000000007, 2.6) {};
\draw (node4) -- (node1);
\node[leaf](node5) at (0.6000000000000001, 2.6) {};
\draw (node5.north) -- (node2);
\node[leaf](node6) at (1.4000000000000001, 2.6) {};
\draw (node6.north) -- (node2);
\node[leaf](node7) at (-2.2, 1.3) {};
\draw (node7.north) -- (node3);
\node[leaf](node8) at (-1.4000000000000001, 1.3) {};
\draw (node8.north) -- (node3);
\node[leaf](node9) at (-0.6000000000000001, 1.3) {};
\draw (node9.north) -- (node4);
\node[leaf](node10) at (0.19999999999999996, 1.3) {};
\draw (node10.north) -- (node4);
\node(node9) at (0, 0.4) {\footnotesize $18 \leq \textrm{VCdim}\,T \leq 38$};
\end{tikzpicture}

\vspace{3mm}

\begin{tikzpicture}[leaf/.style={draw, diamond, minimum width=0.4cm, aspect=.5}, internal/.style={draw, circle, minimum width=0.4cm}, scale=.8]
\node[internal](node0) at (0, 5.2) {};
\node[internal](node1) at (-1.4, 3.9000000000000004) {};
\draw (node1) -- (node0);
\node[internal](node2) at (1.4, 3.9000000000000004) {};
\draw (node2) -- (node0);
\node[internal](node3) at (-2.2, 2.6) {};
\draw (node3) -- (node1);
\node[internal](node4) at (-0.6, 2.6) {};
\draw (node4) -- (node1);
\node[internal](node5) at (1.0, 2.6) {};
\draw (node5) -- (node2);
\node[leaf](node6) at (1.8, 2.6) {};
\draw (node6.north) -- (node2);
\node[leaf](node7) at (-2.6, 1.3) {};
\draw (node7.north) -- (node3);
\node[leaf](node8) at (-1.8, 1.3) {};
\draw (node8.north) -- (node3);
\node[leaf](node9) at (-1.0, 1.3) {};
\draw (node9.north) -- (node4);
\node[leaf](node10) at (-0.19999999999999996, 1.3) {};
\draw (node10.north) -- (node4);
\node[leaf](node11) at (0.6, 1.3) {};
\draw (node11.north) -- (node5);
\node[leaf](node12) at (1.4, 1.3) {};
\draw (node12.north) -- (node5);
\node(node10) at (0, 0.4) {\footnotesize $19 \leq \textrm{VCdim}\,T \leq 47$};
\end{tikzpicture}
\begin{tikzpicture}[leaf/.style={draw, diamond, minimum width=0.4cm, aspect=.5}, internal/.style={draw, circle, minimum width=0.4cm}, scale=.8]
\node[internal](node0) at (0, 5.2) {};
\node[internal](node1) at (-1.6, 3.9000000000000004) {};
\draw (node1) -- (node0);
\node[internal](node2) at (1.6, 3.9000000000000004) {};
\draw (node2) -- (node0);
\node[internal](node3) at (-2.4000000000000004, 2.6) {};
\draw (node3) -- (node1);
\node[internal](node4) at (-0.8, 2.6) {};
\draw (node4) -- (node1);
\node[internal](node5) at (0.8, 2.6) {};
\draw (node5) -- (node2);
\node[internal](node6) at (2.4000000000000004, 2.6) {};
\draw (node6) -- (node2);
\node[leaf](node7) at (-2.8000000000000003, 1.3) {};
\draw (node7.north) -- (node3);
\node[leaf](node8) at (-2.0, 1.3) {};
\draw (node8.north) -- (node3);
\node[leaf](node9) at (-1.2000000000000002, 1.3) {};
\draw (node9.north) -- (node4);
\node[leaf](node10) at (-0.4, 1.3) {};
\draw (node10.north) -- (node4);
\node[leaf](node11) at (0.4, 1.3) {};
\draw (node11.north) -- (node5);
\node[leaf](node12) at (1.2, 1.3) {};
\draw (node12.north) -- (node5);
\node[leaf](node13) at (2.0, 1.3) {};
\draw (node13.north) -- (node6);
\node[leaf](node14) at (2.8000000000000003, 1.3) {};
\draw (node14.north) -- (node6);
\node(node11) at (0, 0.4) {\footnotesize $24 \leq \textrm{VCdim}\,T \leq 52$};
\end{tikzpicture}
\caption{Lower and upper bounds on the VC dimension of the first $11$ non-equivalent trees for $\ell={10}$ real-valued features. Diamond shaped nodes are leaves while circles denote internal nodes.}
\label{figure:vcdim_l=10}
\end{figure}

We applied these algorithms to the first 11 non-equivalent binary decision trees when the number of features is $\ell=10$.
The bounds are presented in Figure~\ref{figure:vcdim_l=10}.
The lower bounds were obtained by the algorithm of Figure~7 of \cite{yildiz2015vc} in conjunction with our exact value for the VC dimension of a decision stump.
Using our base case improves considerably the lower bound found by \cite{yildiz2015vc}.

\clearpage
\section{Supplementary materials about the experiments}
\label{app:experiments}

In this Appendix, we provide more details about the experiments that were done.

\subsection{The pruning algorithm}
\label{app:pruning_algorithm}

The full formal pruning algorithm is given in Algorithm~\ref{algo:prune_tree_with_bound}.

\begin{algorithm}
\caption{PruneTreeWithBound$(t, \epsilon, \delta, m)$}\label{algo:prune_tree_with_bound}
\DontPrintSemicolon
\SetAlgoVlined

\KwIn{A fully grown tree $t$, a bound funtion $\epsilon$ on the true risk, a confidence internal $\delta$, the number of examples $m$.}

Let $T_d$ be the tree class of the tree $t$ with complexity index $d$.

Let $k_t$ be the number of errors made by $t$.

Let $b \leftarrow \epsilon(m,k_t,d,\delta)$ according to Equation~\eqref{eq:shawe-taylor_bound}.

Let $B \leftarrow b$ be the final bound.

\While{$t$ is not a leaf}{
    \For{\textup{every internal node $n$ of the tree $t$}}{
        Let $t_n$ be the tree $t$ with node $n$ replaced by a leaf.
        
        Let $T_{d_n}$ be the tree class of the tree $t_n$ with complexity index $d_n$.

        Let $k_{t_n}$ be the number of errors made by $t_n$.
        
        \If{$\epsilon(m, k_{t_n}, d_n, \delta) \le b$}{
            Let $b \leftarrow \epsilon(m, k_{t_n}, d_n, \delta)$ be the new best bound.
            
            Let $t' \leftarrow t_n$ be the new best tree.
        }
    }
    \uIf{$b \le B$}{
        Let $t \leftarrow t'$.
    }
    \Else{
        \textup{\textbf{break}}
    }
}

\KwOut{The pruned tree $t$, the associated bound $B$.}
\end{algorithm}

\subsection{More statistics about model performances}

We here give more statics on the performances of the model tested, such as the training accuracy, the number of leaves and the height of the final tree, the time it took to prune the original tree, and the computed bound in the case of our pruning algorithm.
For each table, the caption gives the dataset name, the total number of examples it contains, the number of features each example has as well as the number of classes to predict.
For more details about the table columns, see the methodology section~\ref{ssec:methodology}.

All experiments were run on a Intel Core i5-750 CPU running Windows 10, with 12 Go of RAM.

\begin{table}[h!]
\centering
\caption{Breast Cancer Wisconsin Diagnostic Dataset \citep{street1993nuclear} (569 examples, 30 features, 2 classes)}
\vspace{6pt}
\small
\begin{tabular}{ccccc}
\toprule
 & Original & CART & M-CART & Ours\\
\cmidrule{2-5}
Train acc. & $1.000 \pm 0.000$ & $0.962 \pm 0.024$ & $0.965 \pm 0.020$ & $0.983 \pm 0.005$\\
Test acc. & $0.928 \pm 0.024$ & $0.923 \pm 0.027$ & $0.930 \pm 0.017$ & $\mathbf{\mathbf{0.942 \pm 0.022}}$\\
Leaves & $18.0 \pm 2.6$ & $5.9 \pm 3.3$ & $5.8 \pm 3.4$ & $8.3 \pm 1.4$\\
Height & $7.0 \pm 1.0$ & $3.4 \pm 1.6$ & $3.2 \pm 1.4$ & $4.4 \pm 0.6$\\
Time $[s]$ & N/A & $5.3 \pm 0.5$ & $5.3 \pm 0.5$ & $0.1 \pm 0.0$\\
Bound & N/A & N/A & N/A & $1.5 \pm 0.2$\\
\bottomrule
\end{tabular}
\end{table}
\begin{table}[h!]
\centering
\caption{Cardiotocography 10 Dataset \citep{ayres2000sisporto} (2126 examples, 21 features, 10 classes)}
\vspace{6pt}
\small
\begin{tabular}{ccccc}
\toprule
 & Original & CART & M-CART & Ours\\
\cmidrule{2-5}
Train acc. & $0.604 \pm 0.008$ & $0.582 \pm 0.014$ & $0.586 \pm 0.014$ & $0.591 \pm 0.008$\\
Test acc. & $\mathbf{0.566 \pm 0.023}$ & $0.562 \pm 0.023$ & $\mathbf{0.566 \pm 0.024}$ & $\mathbf{\mathbf{0.567 \pm 0.022}}$\\
Leaves & $40.0 \pm 0.0$ & $9.0 \pm 6.2$ & $11.2 \pm 7.0$ & $11.6 \pm 2.7$\\
Height & $15.6 \pm 2.4$ & $5.1 \pm 2.6$ & $5.9 \pm 2.7$ & $6.8 \pm 1.3$\\
Time $[s]$ & N/A & $25.4 \pm 1.5$ & $25.6 \pm 1.6$ & $48.3 \pm 22.0$\\
Bound & N/A & N/A & N/A & $16.8 \pm 0.3$\\
\bottomrule
\end{tabular}
\end{table}
\begin{table}[h!]
\centering
\caption{Climate Model Simulation Crashes Dataset \citep{lucas2013failure} (540 examples, 18 features, 2 classes)}
\vspace{6pt}
\small
\begin{tabular}{ccccc}
\toprule
 & Original & CART & M-CART & Ours\\
\cmidrule{2-5}
Train acc. & $1.000 \pm 0.000$ & $0.918 \pm 0.019$ & $0.941 \pm 0.022$ & $0.977 \pm 0.008$\\
Test acc. & $0.903 \pm 0.024$ & $\mathbf{0.920 \pm 0.021}$ & $\mathbf{\mathbf{0.922 \pm 0.017}}$ & $\mathbf{0.921 \pm 0.014}$\\
Leaves & $21.2 \pm 3.2$ & $1.7 \pm 2.7$ & $3.8 \pm 2.9$ & $9.6 \pm 2.2$\\
Height & $7.2 \pm 1.3$ & $0.5 \pm 1.7$ & $2.4 \pm 1.9$ & $5.2 \pm 0.8$\\
Time $[s]$ & N/A & $4.5 \pm 0.8$ & $4.5 \pm 0.8$ & $0.2 \pm 0.1$\\
Bound & N/A & N/A & N/A & $1.9 \pm 0.2$\\
\bottomrule
\end{tabular}
\end{table}
\begin{table}[h!]
\centering
\caption{Connectionist Bench Sonar Dataset \citep{gorman1988analysis} (208 examples, 60 features, 2 classes)}
\vspace{6pt}
\small
\begin{tabular}{ccccc}
\toprule
 & Original & CART & M-CART & Ours\\
\cmidrule{2-5}
Train acc. & $1.000 \pm 0.000$ & $0.853 \pm 0.117$ & $0.877 \pm 0.120$ & $0.963 \pm 0.012$\\
Test acc. & $\mathbf{\mathbf{0.727 \pm 0.061}}$ & $0.702 \pm 0.054$ & $0.695 \pm 0.084$ & $0.724 \pm 0.053$\\
Leaves & $16.4 \pm 1.7$ & $6.0 \pm 3.3$ & $7.3 \pm 4.4$ & $10.4 \pm 1.6$\\
Height & $6.4 \pm 0.9$ & $3.3 \pm 1.7$ & $3.6 \pm 1.9$ & $5.0 \pm 0.6$\\
Time $[s]$ & N/A & $2.8 \pm 0.3$ & $2.7 \pm 0.2$ & $0.1 \pm 0.1$\\
Bound & N/A & N/A & N/A & $4.5 \pm 0.4$\\
\bottomrule
\end{tabular}
\end{table}
\begin{table}[h!]
\centering
\caption{Diabetic Retinopathy Debrecen Dataset \citep{antal2014ensemble} (1151 examples, 19 features, 2 classes)}
\vspace{6pt}
\small
\begin{tabular}{ccccc}
\toprule
 & Original & CART & M-CART & Ours\\
\cmidrule{2-5}
Train acc. & $0.717 \pm 0.021$ & $0.598 \pm 0.062$ & $0.625 \pm 0.058$ & $0.696 \pm 0.023$\\
Test acc. & $0.613 \pm 0.027$ & $0.576 \pm 0.044$ & $0.602 \pm 0.040$ & $\mathbf{\mathbf{0.622 \pm 0.023}}$\\
Leaves & $40.0 \pm 0.0$ & $2.6 \pm 1.9$ & $4.3 \pm 4.4$ & $16.5 \pm 4.3$\\
Height & $10.7 \pm 1.1$ & $1.5 \pm 1.6$ & $2.4 \pm 2.4$ & $7.8 \pm 1.5$\\
Time $[s]$ & N/A & $10.6 \pm 0.7$ & $10.7 \pm 0.6$ & $2.6 \pm 0.8$\\
Bound & N/A & N/A & N/A & $13.1 \pm 0.8$\\
\bottomrule
\end{tabular}
\end{table}
\begin{table}[h!]
\centering
\caption{Fertility Dataset \citep{gil2012predicting} (100 examples, 9 features, 2 classes)}
\vspace{6pt}
\small
\begin{tabular}{ccccc}
\toprule
 & Original & CART & M-CART & Ours\\
\cmidrule{2-5}
Train acc. & $0.992 \pm 0.007$ & $0.886 \pm 0.025$ & $0.881 \pm 0.017$ & $0.888 \pm 0.027$\\
Test acc. & $0.790 \pm 0.060$ & $\mathbf{\mathbf{0.878 \pm 0.051}}$ & $\mathbf{0.878 \pm 0.051}$ & $0.866 \pm 0.056$\\
Leaves & $14.6 \pm 2.2$ & $1.4 \pm 1.4$ & $1.3 \pm 0.5$ & $1.6 \pm 1.4$\\
Height & $6.9 \pm 1.1$ & $0.4 \pm 1.4$ & $0.3 \pm 0.5$ & $0.5 \pm 1.3$\\
Time $[s]$ & N/A & $0.7 \pm 0.1$ & $0.6 \pm 0.1$ & $0.1 \pm 0.1$\\
Bound & N/A & N/A & N/A & $5.0 \pm 0.7$\\
\bottomrule
\end{tabular}
\end{table}
\begin{table}[h!]
\centering
\caption{Habermans Survival Dataset \citep{haberman1976generalized} (306 examples, 3 features, 2 classes)}
\vspace{6pt}
\small
\begin{tabular}{ccccc}
\toprule
 & Original & CART & M-CART & Ours\\
\cmidrule{2-5}
Train acc. & $0.832 \pm 0.024$ & $0.732 \pm 0.015$ & $0.750 \pm 0.021$ & $0.760 \pm 0.025$\\
Test acc. & $0.660 \pm 0.062$ & $\mathbf{\mathbf{0.746 \pm 0.043}}$ & $0.721 \pm 0.043$ & $0.719 \pm 0.043$\\
Leaves & $40.0 \pm 0.0$ & $1.0 \pm 0.0$ & $3.1 \pm 1.9$ & $3.4 \pm 1.8$\\
Height & $12.4 \pm 1.5$ & $0.0 \pm 0.0$ & $2.0 \pm 1.7$ & $2.1 \pm 1.4$\\
Time $[s]$ & N/A & $4.4 \pm 0.3$ & $4.4 \pm 0.3$ & $1.6 \pm 0.2$\\
Bound & N/A & N/A & N/A & $10.1 \pm 0.8$\\
\bottomrule
\end{tabular}
\end{table}
\begin{table}[h!]
\centering
\caption{Image Segmentation Dataset  (210 examples, 19 features, 7 classes)}
\vspace{6pt}
\small
\begin{tabular}{ccccc}
\toprule
 & Original & CART & M-CART & Ours\\
\cmidrule{2-5}
Train acc. & $1.000 \pm 0.000$ & $0.936 \pm 0.126$ & $0.960 \pm 0.035$ & $0.964 \pm 0.010$\\
Test acc. & $\mathbf{\mathbf{0.862 \pm 0.048}}$ & $0.814 \pm 0.144$ & $0.844 \pm 0.050$ & $0.858 \pm 0.050$\\
Leaves & $17.0 \pm 1.4$ & $10.8 \pm 3.3$ & $11.2 \pm 2.7$ & $10.6 \pm 1.1$\\
Height & $9.8 \pm 1.3$ & $7.0 \pm 1.6$ & $7.4 \pm 1.3$ & $7.4 \pm 1.0$\\
Time $[s]$ & N/A & $2.0 \pm 0.1$ & $2.0 \pm 0.1$ & $8.0 \pm 4.6$\\
Bound & N/A & N/A & N/A & $4.6 \pm 0.3$\\
\bottomrule
\end{tabular}
\end{table}
\begin{table}[h!]
\centering
\caption{Ionosphere Dataset \citep{sigillito1989classification} (351 examples, 34 features, 2 classes)}
\vspace{6pt}
\small
\begin{tabular}{ccccc}
\toprule
 & Original & CART & M-CART & Ours\\
\cmidrule{2-5}
Train acc. & $1.000 \pm 0.000$ & $0.809 \pm 0.132$ & $0.916 \pm 0.051$ & $0.968 \pm 0.009$\\
Test acc. & $\mathbf{0.891 \pm 0.035}$ & $0.772 \pm 0.108$ & $0.867 \pm 0.057$ & $\mathbf{\mathbf{0.892 \pm 0.032}}$\\
Leaves & $19.6 \pm 2.0$ & $3.4 \pm 3.6$ & $5.2 \pm 3.7$ & $9.3 \pm 1.6$\\
Height & $9.6 \pm 2.0$ & $1.9 \pm 2.3$ & $3.2 \pm 2.1$ & $5.4 \pm 0.8$\\
Time $[s]$ & N/A & $4.6 \pm 0.5$ & $4.6 \pm 0.5$ & $0.4 \pm 0.2$\\
Bound & N/A & N/A & N/A & $2.8 \pm 0.2$\\
\bottomrule
\end{tabular}
\end{table}
\begin{table}[h!]
\centering
\caption{Iris Dataset \citep{fisher1936use} (150 examples, 4 features, 3 classes)}
\vspace{6pt}
\small
\begin{tabular}{ccccc}
\toprule
 & Original & CART & M-CART & Ours\\
\cmidrule{2-5}
Train acc. & $1.000 \pm 0.000$ & $0.923 \pm 0.116$ & $0.901 \pm 0.130$ & $0.986 \pm 0.009$\\
Test acc. & $0.933 \pm 0.030$ & $0.860 \pm 0.139$ & $0.838 \pm 0.158$ & $\mathbf{\mathbf{0.937 \pm 0.028}}$\\
Leaves & $7.6 \pm 1.3$ & $3.9 \pm 1.5$ & $3.8 \pm 1.4$ & $4.8 \pm 1.0$\\
Height & $4.8 \pm 0.8$ & $2.8 \pm 1.4$ & $2.8 \pm 1.4$ & $3.6 \pm 0.8$\\
Time $[s]$ & N/A & $0.5 \pm 0.1$ & $0.6 \pm 0.1$ & $0.0 \pm 0.0$\\
Bound & N/A & N/A & N/A & $2.0 \pm 0.3$\\
\bottomrule
\end{tabular}
\end{table}
\begin{table}[h!]
\centering
\caption{Parkinson Dataset \citep{little2007exploiting} (195 examples, 22 features, 2 classes)}
\vspace{6pt}
\small
\begin{tabular}{ccccc}
\toprule
 & Original & CART & M-CART & Ours\\
\cmidrule{2-5}
Train acc. & $1.000 \pm 0.000$ & $0.908 \pm 0.098$ & $0.944 \pm 0.060$ & $0.976 \pm 0.013$\\
Test acc. & $0.859 \pm 0.062$ & $0.848 \pm 0.064$ & $0.858 \pm 0.065$ & $\mathbf{\mathbf{0.863 \pm 0.065}}$\\
Leaves & $12.7 \pm 1.8$ & $5.6 \pm 3.5$ & $6.8 \pm 3.4$ & $8.2 \pm 1.3$\\
Height & $5.7 \pm 1.1$ & $3.0 \pm 2.0$ & $3.6 \pm 1.6$ & $4.0 \pm 0.7$\\
Time $[s]$ & N/A & $1.4 \pm 0.1$ & $1.4 \pm 0.1$ & $0.0 \pm 0.0$\\
Bound & N/A & N/A & N/A & $3.1 \pm 0.4$\\
\bottomrule
\end{tabular}
\end{table}
\begin{table}[h!]
\centering
\caption{Planning Relax Dataset \citep{bhatt2012planning} (182 examples, 12 features, 2 classes)}
\vspace{6pt}
\small
\begin{tabular}{ccccc}
\toprule
 & Original & CART & M-CART & Ours\\
\cmidrule{2-5}
Train acc. & $1.000 \pm 0.000$ & $0.720 \pm 0.038$ & $0.709 \pm 0.016$ & $1.000 \pm 0.000$\\
Test acc. & $0.595 \pm 0.075$ & $0.725 \pm 0.049$ & $\mathbf{\mathbf{0.729 \pm 0.048}}$ & $0.595 \pm 0.075$\\
Leaves & $29.1 \pm 2.2$ & $1.7 \pm 2.6$ & $1.0 \pm 0.0$ & $29.1 \pm 2.2$\\
Height & $11.4 \pm 2.1$ & $0.6 \pm 2.3$ & $0.0 \pm 0.0$ & $11.4 \pm 2.1$\\
Time $[s]$ & N/A & $2.7 \pm 0.3$ & $2.0 \pm 0.2$ & $1.0 \pm 0.3$\\
Bound & N/A & N/A & N/A & $6.5 \pm 0.1$\\
\bottomrule
\end{tabular}
\end{table}
\begin{table}[h!]
\centering
\caption{QSAR Biodegradation Dataset \citep{mansouri2013quantitative} (1055 examples, 41 features, 2 classes)}
\vspace{6pt}
\small
\begin{tabular}{ccccc}
\toprule
 & Original & CART & M-CART & Ours\\
\cmidrule{2-5}
Train acc. & $0.834 \pm 0.022$ & $0.758 \pm 0.042$ & $0.791 \pm 0.026$ & $0.804 \pm 0.025$\\
Test acc. & $0.752 \pm 0.031$ & $0.741 \pm 0.033$ & $0.757 \pm 0.026$ & $\mathbf{\mathbf{0.761 \pm 0.028}}$\\
Leaves & $40.0 \pm 0.0$ & $3.1 \pm 2.9$ & $7.0 \pm 4.8$ & $10.3 \pm 4.0$\\
Height & $12.8 \pm 1.7$ & $1.8 \pm 1.9$ & $4.3 \pm 2.2$ & $5.7 \pm 1.8$\\
Time $[s]$ & N/A & $16.7 \pm 1.9$ & $16.2 \pm 1.0$ & $2.9 \pm 0.8$\\
Bound & N/A & N/A & N/A & $8.5 \pm 0.8$\\
\bottomrule
\end{tabular}
\end{table}
\begin{table}[h!]
\centering
\caption{Seeds Dataset \citep{charytanowicz2010complete} (210 examples, 7 features, 3 classes)}
\vspace{6pt}
\small
\begin{tabular}{ccccc}
\toprule
 & Original & CART & M-CART & Ours\\
\cmidrule{2-5}
Train acc. & $1.000 \pm 0.000$ & $0.967 \pm 0.019$ & $0.964 \pm 0.057$ & $0.981 \pm 0.007$\\
Test acc. & $0.918 \pm 0.034$ & $0.914 \pm 0.040$ & $0.905 \pm 0.081$ & $\mathbf{\mathbf{0.925 \pm 0.033}}$\\
Leaves & $12.0 \pm 1.8$ & $5.7 \pm 1.8$ & $6.4 \pm 2.1$ & $7.0 \pm 0.9$\\
Height & $6.0 \pm 0.9$ & $3.9 \pm 1.1$ & $4.1 \pm 1.1$ & $4.3 \pm 0.5$\\
Time $[s]$ & N/A & $1.1 \pm 0.1$ & $1.2 \pm 0.1$ & $0.1 \pm 0.1$\\
Bound & N/A & N/A & N/A & $2.4 \pm 0.4$\\
\bottomrule
\end{tabular}
\end{table}
\begin{table}[h!]
\centering
\caption{Spambase Dataset  (4601 examples, 57 features, 2 classes)}
\vspace{6pt}
\small
\begin{tabular}{ccccc}
\toprule
 & Original & CART & M-CART & Ours\\
\cmidrule{2-5}
Train acc. & $0.861 \pm 0.026$ & $0.846 \pm 0.026$ & $0.850 \pm 0.028$ & $0.855 \pm 0.026$\\
Test acc. & $0.844 \pm 0.027$ & $0.839 \pm 0.028$ & $0.842 \pm 0.029$ & $\mathbf{\mathbf{0.846 \pm 0.026}}$\\
Leaves & $40.0 \pm 0.0$ & $7.0 \pm 5.6$ & $8.1 \pm 5.6$ & $9.6 \pm 4.0$\\
Height & $19.8 \pm 2.4$ & $3.8 \pm 2.5$ & $4.4 \pm 2.8$ & $5.5 \pm 2.0$\\
Time $[s]$ & N/A & $82.4 \pm 10.1$ & $81.6 \pm 10.6$ & $3.4 \pm 0.4$\\
Bound & N/A & N/A & N/A & $6.0 \pm 1.0$\\
\bottomrule
\end{tabular}
\end{table}
\begin{table}[h!]
\centering
\caption{Vertebral Column 3C Dataset \citep{berthonnaud2005analysis} (310 examples, 6 features, 3 classes)}
\vspace{6pt}
\small
\begin{tabular}{ccccc}
\toprule
 & Original & CART & M-CART & Ours\\
\cmidrule{2-5}
Train acc. & $1.000 \pm 0.000$ & $0.784 \pm 0.181$ & $0.881 \pm 0.044$ & $0.952 \pm 0.019$\\
Test acc. & $0.800 \pm 0.050$ & $0.725 \pm 0.139$ & $0.804 \pm 0.046$ & $\mathbf{\mathbf{0.819 \pm 0.044}}$\\
Leaves & $31.8 \pm 3.4$ & $6.5 \pm 7.0$ & $6.4 \pm 4.2$ & $15.6 \pm 3.2$\\
Height & $9.8 \pm 1.3$ & $3.5 \pm 3.5$ & $4.3 \pm 2.2$ & $7.7 \pm 1.7$\\
Time $[s]$ & N/A & $3.0 \pm 0.3$ & $2.9 \pm 0.3$ & $12.5 \pm 3.4$\\
Bound & N/A & N/A & N/A & $4.6 \pm 0.4$\\
\bottomrule
\end{tabular}
\end{table}
\begin{table}[h!]
\centering
\caption{Wall Following Robot 24 Dataset \citep{freire2009short} (5456 examples, 24 features, 4 classes)}
\vspace{6pt}
\small
\begin{tabular}{ccccc}
\toprule
 & Original & CART & M-CART & Ours\\
\cmidrule{2-5}
Train acc. & $1.000 \pm 0.000$ & $0.999 \pm 0.001$ & $0.999 \pm 0.001$ & $0.998 \pm 0.001$\\
Test acc. & $\mathbf{\mathbf{0.995 \pm 0.002}}$ & $\mathbf{0.994 \pm 0.002}$ & $\mathbf{0.994 \pm 0.002}$ & $\mathbf{0.994 \pm 0.001}$\\
Leaves & $28.5 \pm 3.3$ & $22.3 \pm 4.8$ & $22.8 \pm 4.5$ & $17.8 \pm 1.5$\\
Height & $9.6 \pm 1.0$ & $9.3 \pm 1.3$ & $9.3 \pm 1.3$ & $8.6 \pm 1.0$\\
Time $[s]$ & N/A & $59.4 \pm 3.0$ & $57.6 \pm 3.1$ & $32.9 \pm 20.5$\\
Bound & N/A & N/A & N/A & $0.3 \pm 0.0$\\
\bottomrule
\end{tabular}
\end{table}
\begin{table}[h!]
\centering
\caption{Wine Dataset \citep{aeberhard1994comparative} (178 examples, 13 features, 3 classes)}
\vspace{6pt}
\small
\begin{tabular}{ccccc}
\toprule
 & Original & CART & M-CART & Ours\\
\cmidrule{2-5}
Train acc. & $1.000 \pm 0.000$ & $0.981 \pm 0.015$ & $0.984 \pm 0.012$ & $0.989 \pm 0.010$\\
Test acc. & $\mathbf{\mathbf{0.908 \pm 0.041}}$ & $0.902 \pm 0.045$ & $0.903 \pm 0.043$ & $0.904 \pm 0.046$\\
Leaves & $8.0 \pm 2.3$ & $5.6 \pm 1.6$ & $5.9 \pm 1.8$ & $6.3 \pm 1.2$\\
Height & $4.2 \pm 1.2$ & $3.2 \pm 0.6$ & $3.4 \pm 0.8$ & $3.2 \pm 0.4$\\
Time $[s]$ & N/A & $0.8 \pm 0.1$ & $0.8 \pm 0.1$ & $0.0 \pm 0.0$\\
Bound & N/A & N/A & N/A & $2.2 \pm 0.6$\\
\bottomrule
\end{tabular}
\end{table}
\begin{table}[h!]
\centering
\caption{Yeast Dataset \citep{horton1996probabilistic} (1484 examples, 8 features, 10 classes)}
\vspace{6pt}
\small
\begin{tabular}{ccccc}
\toprule
 & Original & CART & M-CART & Ours\\
\cmidrule{2-5}
Train acc. & $0.470 \pm 0.007$ & $0.370 \pm 0.057$ & $0.386 \pm 0.058$ & $0.449 \pm 0.008$\\
Test acc. & $0.429 \pm 0.019$ & $0.368 \pm 0.059$ & $0.384 \pm 0.058$ & $\mathbf{\mathbf{0.442 \pm 0.019}}$\\
Leaves & $40.0 \pm 0.0$ & $2.0 \pm 1.1$ & $2.7 \pm 2.0$ & $6.2 \pm 1.3$\\
Height & $14.2 \pm 2.0$ & $1.0 \pm 1.1$ & $1.6 \pm 1.7$ & $4.1 \pm 0.9$\\
Time $[s]$ & N/A & $8.5 \pm 0.4$ & $8.4 \pm 0.3$ & $418.9 \pm 74.5$\\
Bound & N/A & N/A & N/A & $22.3 \pm 0.3$\\
\bottomrule
\end{tabular}
\end{table}

\end{document}